\documentclass[10pt,twocolumn,letterpaper]{article}

\usepackage{iccv}
\usepackage{times}
\usepackage{epsfig}
\usepackage{graphicx}
\usepackage{amsmath}
\usepackage{amssymb}

\usepackage{graphicx}
\usepackage{amsthm}
\usepackage{mathrsfs} %花体字母
\usepackage{amsmath}
\usepackage{amssymb}
\usepackage{booktabs}
\usepackage{multirow} % for multicolumn and multirow table
\usepackage{colortbl}
\definecolor{mygray}{gray}{.9}
\usepackage{bm}
\usepackage[pagebackref=true,breaklinks=true,letterpaper=true,colorlinks,bookmarks=false]{hyperref}

\usepackage[capitalize]{cleveref}
\crefname{section}{Sec.}{Secs.}
\Crefname{section}{Section}{Sections}
\Crefname{table}{Table}{Tables}
\crefname{table}{Tab.}{Tabs.}
\Crefname{figure}{Figure}{Figures}
\crefname{figure}{Fig.}{Figs.}

\usepackage{authblk}

\iccvfinalcopy % *** Uncomment this line for the final submission

\ificcvfinal\fi

\vspace{-1.0em}
\title{PGformer: Proxy-Bridged Game Transformer for Multi-Person \\ Highly Interactive Extreme Motion Prediction}

\newcommand*{\affaddr}[1]{#1} % No op here. Customize it for different styles.
\newcommand*{\affmark}[1][*]{\textsuperscript{#1}}
\newcommand*{\email}[1]{\texttt{#1}}

\author{%
Yanwen Fang\affmark[1,2], 
Jintai Chen$^{\dagger}$\affmark[3], 
Peng-Tao Jiang\affmark[3], 
Chao Li\affmark[2], 
\\
Yifeng Geng\affmark[2], 
Eddy K.F. LAM\affmark[1], 
Guodong Li$^{\dagger}$\affmark[1] \\
\affaddr{\affmark[1]The University of Hong Kong},
\affaddr{\affmark[2]DAMO Academy, Alibaba}, 
\affaddr{\affmark[3]Zhejiang University} \\
\small
 \email{u3545683@connect.hku.hk, jtigerchen@zju.edu.cn, pt.jiang@mail.nankai.edu.cn, lllcho.lc@alibaba-inc.com, gengyifeng@gmail.com, hrntlkf@hku.hk, gdli@hku.hk} \\
}
\newcommand\nnfootnote[1]{%
  \begin{NoHyper}
  \renewcommand\thefootnote{}\footnote{#1}%
  \addtocounter{footnote}{-1}%
  \end{NoHyper}
}

\begin{document}

\maketitle
\nnfootnote{$\dagger$ Corresponding authors.}
\ificcvfinal\thispagestyle{empty}\fi

\vspace{-2em}
%%%%%%%%% ABSTRACT
\begin{abstract}
Multi-person motion prediction is a challenging task, especially for real-world scenarios of highly interacted persons. 
Most previous works have been devoted to studying the case of weak interactions (e.g., walking together), in which typically forecasting each human pose in isolation can still achieve good performances. 
This paper focuses on collaborative motion prediction for multiple persons with extreme motions and attempts to explore the relationships between the highly interactive persons' pose trajectories. 
Specifically, a novel cross-query attention (XQA) module is proposed to bilaterally learn the cross-dependencies between the two pose sequences tailored for this situation. 
A proxy unit is additionally introduced to bridge the involved persons, which cooperates with our proposed XQA module and subtly controls the bidirectional spatial information flows. 
% acting as a motion intermediary. 
These designs are then integrated into a Transformer-based architecture and the resulting model is called \textbf{P}roxy-bridged \textbf{G}ame Trans\textbf{former} (\textbf{PGformer}) for multi-person interactive motion prediction. 
Its effectiveness has been evaluated on the challenging ExPI dataset, which involves highly interactive actions. 
Our PGformer consistently outperforms the state-of-the-art methods in both short- and long-term predictions by a large margin. 
Besides, our approach can also be compatible with the weakly interacted CMU-Mocap and MuPoTS-3D datasets and extended to the case of more than 2 individuals with encouraging results. 
Our code will be publicly available upon acceptance. 
\end{abstract}

%%%%%%%%% BODY TEXT
%%%%%%%%%%%%%%%%%%%%%%%%%%%%%%%%%%%%%%%%%%
\section{Introduction} 
~\label{sec:intro}
% human motion prediction
Human motion prediction aims to forecast a sequence of future 3D motion trajectories given a sequence of past ones, 
% which has many real-world computer vision applications, 
which is widely used in autonomous driving~\cite{djuric2020uncertainty,akbari2017automatic}, target tracking~\cite{6126296}, and human-robot interaction~\cite{Koppula2013AnticipatingHA, butepage2018anticipating}, \textit{et al}.
% surveillance systems. 
The skeleton-based human motion sequence is a structured time series, namely, the movement of a single body joint is affected by the coupling of spatial connections with other joints and the temporal trajectory tendency.
Therefore, most existing works reformulated motion prediction as a sequence-to-sequence prediction task and tackled this problem by deep learning models~\cite{julieta2017motion, Aksan_2019_ICCV, Martinez_potr_ICCV2021, 9665904, Zhong2022SpatioTemporalGG, li2022spgsn}. 
% were developed to extract these sequence properties from past poses.
% with the recent popularity of deep learning models like RNN, Transformer and GCN~\cite{julieta2017motion, Aksan_2019_ICCV, Martinez_potr_ICCV2021, 9665904, Zhong2022SpatioTemporalGG, sofianos2021spacetimeseparable}.
% past observations of 3D skeleton data are used to forecast future skeleton movements.
%%%%%%%%%%%%%%%%%%%%%%%%%%%%%%
% why multi-person
% learning person interaction is important
While previous works have achieved remarkable successes, most of them are devoted to exploring single-person prediction which forecasts single human poses in isolation, limiting their direct applications in real-world scenarios of multiple persons. 
% essentially ignoring the interactions between humans, even in the case of multi-person motion prediction. 
Growing evidence has shown that the motion of one person is typically affected by those of others in a scenario with multiple persons~\cite{wang2021multiperson}. 
% tackled this problem as a single human pose forecasting in isolation
% In real-world scenarios of multiple persons, people continuously interact with each other, whose actions and behaviors may highly depend on the other persons around.
% Hence, 
Recently, multi-person motion prediction has drawn increasing attention from researchers~\cite{wang2021multiperson, Vendrow2022SoMoFormer, peng2023trajectory}. 
% which has broader real-world applications, 
% and they have attained some achievements by integrating interacted persons' information into the motion prediction for one person in a straightforward way (e.g., by concatenation)~\cite{wang2021multiperson, Vendrow2022SoMoFormer, Adeli2021TRiPOD}.
% which verified the necessity of learning information from the interacted people  % or supporting 
Even so, these works mainly focused on modeling weakly interacted persons (e.g., hand-shaking and walking together), and yet neglected the scenario of highly correlated persons that is often seen in team sports or collaborative assembly tasks. 

\begin{figure}[t]
    \begin{center}
        \centerline{\includegraphics[width=1.0\linewidth]{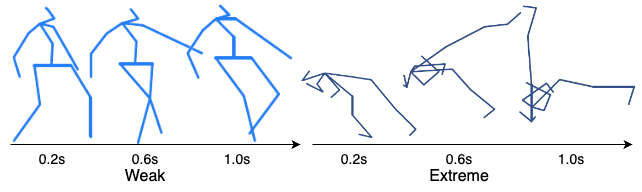}} % \linewidth
        % \vskip -0.1in
        \caption{Weakly interacted motions vs highly interacted extreme motions.}
        \label{fig:comp_extreme}
    \end{center}
    \vskip -0.4in
\end{figure}

% \begin{table}[t]
%     % \setlength\tabcolsep{2.8pt}
%     % \vskip -0.05in
%     \caption{Results of the compared methods on ExPI. }
%     % \vskip -0.2in
%     \label{tab:motivation}
%     \begin{center}
%     % \footnotesize
%     \small
%     \begin{tabular}{l|cccc}
%         \hline
%         % \toprule
%         Time (sec) & 0.2 & 0.4 & 0.6 & 1.0 \\
%         \hline
%         % \midrule
%         \cite{wang2021multiperson} \\
%         \cite{peng2023trajectory} \\
%         PGformer \\

%         \hline
%         % \bottomrule
%     \end{tabular}
%     \end{center}
%     \vskip -0.3in
% \end{table}

% Hence encouraging cross multiple
% guo wen's paper
To fill this gap, 
% in this sub-research area, 
Guo \textit{et al.}~\cite{guo2021multi} collected a new dataset called ExPI (Extreme Pose Interaction) containing human motion sequences in pair with extreme interactions. 
% with two persons performing highly interactive activities.
Based on this dataset, they also proposed a   cross-interaction attention (XIA) module and used it before a 26-layer GCN, which is the first work to employ the historical information of both people in an interactive fashion. 
However, their model only exploited the multi-person interactions in the historical poses, predicting the future poses individually without any cross-interaction. 
% of these two persons 
% is elaborately designed for this dataset, for example, it applies XIA modules for chunked key and value of each person respectively, limiting the applications to other datasets. 
% Besides, the XIA module is only applied in the encoder and predicts the future pose sequences of these two persons individually without interactions.
% Besides, the overall framework focuses more on learning spatial dependencies between the different body parts via a 26-layer GCN.
To clarify the word `extreme', \Cref{fig:comp_extreme} compares weakly interacted motions and highly interacted extreme motions, showing that the right ones are more complex.  
% we give the statistics in appendix
% \Cref{fig:comp_extreme} (b) shows that the works~\cite{wang2021multiperson, peng2023trajectory} exploring weak interactions are not well generalized on extreme motions. 

% in this paper, we propose
This paper attempts to study the interactions of multiple persons' pose trajectories with extreme motions from the past to the future, and proposes an end-to-end Transformer-based network for multi-person pose forecasting. 
% Our approach is simple yet effective, and its overall pipeline is illustrated briefly in \Cref{fig:framework}. 
Specifically, a novel cross-query attention (XQA) module is first proposed to bilaterally learn the cross-dependencies between the two pose sequences. 
In our XQA, the two persons' poses act as queries in computing attention scores to retrieve useful information for each other and share the same attention map.  
Different from XIA~\cite{guo2021multi}, ours can explicitly handle the entire scenario with multiple interacted persons in one step instead of employing two individual attention modules for each person respectively. 
We use our XQA to learn the cross-interactions between the involved persons' poses from the past to the future.

% To better model the scenario, we further introduce a concept of \textit{proxy} to  
% build an unit to bridge the involved persons, 
% since we assume there exists a \textit{proxy} unit for extreme actions and this concept is motivated by the following. 
% For instance, in a fencing game, the \textit{proxy} is typically the swords that closely affect the two opponents. 
% We consider the games without a \textit{proxy} as a special case (e.g., the boxing game) and can assume a virtual one for this case.
To subtly recalibrate the spatial dependencies of the body joints, we further introduce a concept of \textit{proxy} which includes spatial semantic information and interacts with the involved persons in the entire sequence. 
% acting as a motion intermediary between them. 
% The propose \textit{proxy} leverages 
In addition, we devise two \textit{proxy} units for the past and future poses respectively, and the future one is affected by the past one as well as the future poses in multi-person interactive actions or games. 
In this way, \textit{proxy} acts as a motion intermediary gathering spatial semantic information entangled in body joints and provides a subtle control of the bidirectional spatial information flows. 
Cooperating with our XQA, \textit{proxy} facilitates transferring the effective pose information bilaterally between the involved persons from the past to the future, just like a bridge, and thus the future interactions are well modeled. 
% Based on these two designs, our PGformer 
% multi-person interactive motion prediction.
% We then incorporate these two designs into a Transformer-based architecture and devise a new framework for multi-person interactive motion prediction.
Since this paper builds a Transformer to model the multi-person game (scenario) leveraging an implicitly learnable \textit{proxy}, we call our approach \textit{\textbf{P}roxy-bridged \textbf{G}ame Trans\textbf{former}} (\textbf{PGformer}). 
% Since the two persons shared the same score matrix in the XQA module and interact with each other, just like playing a game, we name our framework Proxy-Bridged Game Transformer (PGformer).
% In the scenarios of extreme human motions,
Besides, gravity loss for each person is introduced into the loss function, which ensures the center of gravity is kept within a plausible altitude range and avoids a large variance of different models. 
% dramatic change in contiguous frames. 

Compared with the elaborately designed model of XIA, our PGformer's architecture stands much closer to currently existing state-of-the-art Transformer-based models, which is much easier to be extended or adapted to other datasets and methods. 
For instance, there is no need to split different chunks for keys and values separately (like the method in~\cite{guo2021multi}) since PGformer can deal with input sequences of arbitrary length. 
We evaluate PGformer on the highly interactive ExPI dataset, and experiments show that our model consistently outperforms the state-of-the-art methods both in short- and long-term predictions.  
% by at least 4--6\% in terms of percentage of improvement on average JME and AME.
We also show that our approach can be compatible with the weakly interacted CMU-Mocap and MuPoTS-3D datasets and extended to more individuals with encouraging results, validating the generalization ability of our model. 
% Our main contributions
% The main contributions of this paper are summarized below: 
To summarize: 
\begin{itemize}
    \item A novel cross-query attention module is proposed to bilaterally learn the cross-dependencies between the two interactive poses, sharing the same attention map. 
    
    \item A concept of \textit{proxy} is additionally introduced, which cooperates with our XQA to subtly recalibrate the effective pose information in a bidirectional manner. 
    
    % model the interactions of the involved persons from the past to the future in our architecture, 
    % better transferring the effective pose information in a bidirectional fashion. 
    % proxy to control the bidirectional information flows like a motion intermediary.
    % 
    \item We propose a framework called PGformer with the above designs for multi-person highly interactive extreme motion prediction, considering the interactions between the involved persons not only in the historical poses but also in predicting future motions. 
    % We incorporate the proposed cross-query attention module with proxy into a Transformer architecture, and devise a new framework called proxy-bridged game Transformer (PGformer) for multi-person highly interactive motion prediction. 
    % 
    \item Experiments show that PGformer surpasses the current state-of-the-art methods in both short- and long-term predictions by at least 4--6\% on ExPI. 
    We also verify that ours can be compatible with the weakly interacted CMU-Mocap and MuPoTS-3D with more persons.  
    % datasets, and can achieve encouraging results on them.
    % \item We investigate the important design components of our proposed attention module and framework, and show the qualitative results to further support our claims.
\end{itemize}

%-------------------------------------------------------------------------
\section{Related Work}
\paragraph{Human Motion Prediction.}
% Human motion prediction has been widely studied in the early period with Recurrent Neural Networks (RNNs) due to the inherent sequential structure of human motion. 
% RNN
% For example, an Encoder-Recurrent-Decoder (ERD) model~\cite{fragkiadaki2015recurrent} incorporated nonlinear encoder and decoder networks before and after LSTM layers.
% The approach prevents catastrophic drift via a schedule to add Gaussian noise to the inputs and increase the model robustness, but this scheduling is hard to tune in practice.
% The work presented in~\cite{julieta2017motion} applied an encoder-decoder RNN architecture with a single GRU unit, and added a residual connection between decoder inputs and outputs as a way of modeling velocities in the predicted sequence. 
% Though RNNs had achieved great success in motion prediction, they suffered from converging to a static pose, which is a general problem met by the auto-regressive method.
% Some works tried to alleviate this problem by convolutional models~\cite{li2018convolutional} and adversarial training~\cite{gui2018adversarial}.
% containing the entire history with a fixed-size hidden state and 
%%%%%%%%%%%%%%%%%%%%%%%%%%%%%%%%%%%%%%%
% GCN 
% by defining a graph with joints as vertices and their natural connections as edges
Since human motion prediction (HMP) is a task of spatio-temporal forecasting, recently some researchers used GCN with trainable adjacency matrices to model the pair-wise joint dependencies of human motion~\cite{mao2019learning}.
Dang \textit{et al.}~\cite{dang2021msr} further developed GCN-based methods by leveraging multi-scale supervision. 
Some works employed spatial GCNs and temporal GCNs or spatio-temporal ones to tackle this task~\cite{sofianos2021spacetimeseparable, Zhong2022SpatioTemporalGG}. 
% Transformer
% Motivated by the success of Transformer~\cite{vaswani2017attention} in Natural Language Processing (NLP), many researchers have applied Transformer-based architectures to motion prediction domains.
Subsequent methods built upon the success of Transformer-based or attention-based models~\cite{vaswani2017attention} for long-term motion prediction.
For instance, Aksan \textit{et al.}~\cite{9665904} adopted spatial attention and temporal attention separately.
Mao \textit{et al.}~\cite{mao2020history} introduced an attention module before a GCN  to capture the similarities between current and historical motion, allowing the model to aggregate past motions for long-term prediction. 
A non-autoregressive Transformer, supervised by an activity classifier, was leveraged to infer the pose sequences in parallel, avoiding error accumulation~\cite{Martinez_potr_ICCV2021}.

% \noindent\textbf{Multi-Person HMP.~~} 
\paragraph{Multi-Person HMP.} 
% \paragraph{Multi-Person Human Motion Prediction.}
One key fact is that humans never live in isolation, they continuously interact with other people and objects in real-world scenarios, which means the motion of one person is typically dependent on or correlated with the motion of other people or objects around~\cite{wang2021multiperson, guo2021multi}. 
% Some previous works devoted to exploring the human-to-object correlations but without human-to-human interactions~\cite{corona2020context, caoHMP2020}.
With the growing need of scenes with multiple people, some recent works have begun to focus on modeling human-to-human interactions.
% forecasting future 3D poses for multi-person in an entire scene. 
%~\cite{Alahi2014Socially} present an LSTM model which jointly reasons across multiple individuals in a scene.
Mohamed \textit{et al.}~\cite{mohamed2020social} employed a graph-based spatio-temporal model called Social-STGCNN with a specific kernel function in the weighted adjacency matrix to learn the social interactions between pedestrians.
TRiPOD~\cite{Adeli2021TRiPOD} used graph attentional networks to model interactions between the involved persons and objects, but applied an RNN with an attentional graph to predict future motion. 
Wang \textit{et al.}~\cite{wang2021multiperson} introduced a Transformer-based architecture assisted by a motion discriminator (MRT), and predicted the motion for one person by concatenating other persons' information in the decoder, which is a straightforward way. 
TBIFormer~\cite{peng2023trajectory} utilized a Social Body Interaction Self-Attention (SBI-MSA) module to learn body part dynamics for inter- and intraindividual interactions. 
% can only make inferences for one person at a time.
Though the works mentioned above aimed at exploring the human-to-human interactions, they only studied the scenes of multiple people with weak interactions and small movements, e.g., moving, standing and chatting.

More recently, Guo \textit{et al.}~\cite{guo2021multi} collected the ExPI dataset and proposed a cross-interaction attention (XIA) mechanism to exploit the historical information of both persons.
Nonetheless, only the encoder utilized the XIA module, but the interactions between the future predicted sequences are neglected.
Rahman \textit{et al.}~\cite{rahman2023best} concatenated the two involved persons together and sent the features into a spatio-temporal GCN with their proposed initialized weights.
However, their model suffered from the problem of instable training and the results had large variances. 
% The elaborately designed framework also 
Besides, the number of persons should be fixed for the model which influences parameters, limiting its broader applications to other datasets with different persons in various scenes. 
It is worth noting that we have tested the model in~\cite{wang2021multiperson, peng2023trajectory} on ExPI in our experiments, and the performances suggest that they are insufficient to learn the cross-dependencies between two highly interactive persons.

%-------------------------------------------------------------------------

\section{Proposed Method}\label{sec:method}
This section first formulates the problem of multi-person human motion prediction. 
Then, it proposes a cross-query attention (XQA) module and a \textit{proxy} unit.
Lastly, it gives the overall network architecture of our approach. 

\subsection{Problem Formulation} 
Given a scene with a pair of persons and their corresponding history motion sequences, our goal is to predict their future 3D motion sequences. 
Similar to~\cite{guo2021multi}, here we use $l$ and $f$ to denote the leader and the follower of two persons to differentiate them. 
Specifically, given two pose sequences $\bm{X}^{l}_{1:T}=[x^{l}_1, \dots, x^{l}_T]$ and $\bm{X}^{f}_{1:T}=[x^{f}_1, \dots, x^{f}_T]$ representing history 3D poses with $T$ time steps, we aim to predict the future $K$ poses $\bm{X}^{l}_{T+1:T+K}$ and $\bm{X}^{f}_{T+1:T+K}$. 
A vector $x^{l}_t \in \mathbb{R}^{3J}$ containing the Cartesian coordinates of the $J$ skeleton joints is used to represent the pose of the person $l$ at time step $t$. 
This forecasting problem is strongly related to conditional sequence modeling where the goal is to model
the joint probability $P(\bm{X}^{l}_{T+1:T+K}, \bm{X}^{f}_{T+1:T+K}|\bm{X}^{l}_{1:T}, \bm{X}^{f}_{1:T}; \theta)$ with model parameters $\theta$. 
In our work, $\theta$ are the parameters of our PGformer.
For simplicity, we omit superscript $l$ or $f$ when the superscript variable only represents an arbitrary person, e.g., taking $\bm{X}^{l}_{1:T}$ or $\bm{X}^{f}_{1:T}$ as $\bm{X}_{1:T}$.

\subsection{Cross-Query Attention Module}
~\label{subsec:xqa_module}
\begin{figure}[t]
    \vskip -0.05in
    \begin{center}
        \centerline{\includegraphics[width=0.9\linewidth]{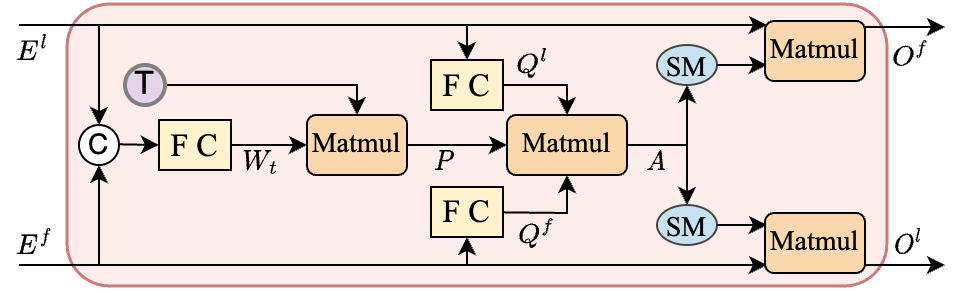}} % \linewidth
        \vskip +0.05in
        \caption{Illustrations of our cross-query attention (XQA) module with a \textit{proxy}, where `SM' and `Matmul' indicate $Softmax$ and matrix multiplication. \textcircled{c} denotes channel-wise concatenation. }
        \label{fig:XQA_module}
    \end{center}
    \vskip -0.4in
\end{figure}

Since our goal is to learn two person-specific motion prediction mappings, we propose a cross-query attention (XQA) module to learn the correlations between these two mappings.
% Our motivation is that the pose information of one person can be used to transform the pose information of the other person for better learning motion properties.
Our motivation is that the pose trajectory of one person can influence the pose trajectory of the other person since they are highly interactive with each other.
The two persons' pose information should be considered simultaneously for better learning motion properties.
Inspired by this, we suppose that the two persons act as queries in computing attention scores to retrieve useful information for each other and share the same attention map. 
The detailed inner elements of our bidirectional XQA module are given in \Cref{fig:XQA_module} and can be illustrated as the following. 

We denote the output of a Transformer layer (the input of an XQA module) as $\bm{E}^l$ and $\bm{E}^f \in \mathbb{R}^{T \times D}$ respectively. 
Here we omit the subscript for simplicity since the shapes of matrices are all given. 
Then the queries are given by:
\vskip -0.2in
\begin{equation} \label{eq:get_queries}
    \bm{Q}^l = \text{ReLU}(\text{FC}(\bm{E}^l)) \hspace{2mm}\text{and}\hspace{2mm} \bm{Q}^f = \text{ReLU}(\text{FC}(\bm{E}^f)), 
\end{equation}
\vskip -0.05in
\noindent{where $\bm{Q}^l$ and $\bm{Q}^f \in \mathbb{R}^{T \times D}$ are the queries for the two persons' poses with the model dimension of $D$, respectively.}
The shared attention score is built by retrieving query-related information from the other query: 
\vskip -0.1in
\begin{equation}\label{eq:shared_attn}
    \bm{A} = \bm{Q}^l[\bm{Q}^f]^{\mathsf{T}}, 
\end{equation}
\vskip -0.05in
\noindent{where $\bm{A} \in \mathbb{R}^{T \times T}$ is the attention map shared by the two interactive persons.}
As the two persons share the same attention map, we apply the $Softmax$ (SM) function along the two different dimensions to obtain the score matrices for them. 
We utilize the inputs of our XQA module as values for attention directly.  
In this way, the final outputs are obtained after reweighting the values by the attention score:
\vskip -0.2in
\begin{equation}\label{eq:sm_attn}
    \bm{O}^l = \text{SM}(\bm{A}) \cdot \bm{E}^f \hspace{3mm}\text{and}\hspace{3mm}  \bm{O}^f = \text{SM}(\bm{A}^{\mathsf{T}}) \cdot \bm{E}^l,
\end{equation}
\vskip -0.1in
\noindent{where $\bm{O}^l$ and $\bm{O}^f$ are the outputs of XQA module.}
Our proposed attention mechanism can be directly extended to a multi-head version. 
We use $\bm{O}^l, \bm{O}^f = \text{XQA}(\bm{E}^l, \bm{E}^f)$ to summarize the above equations (Eq.~(\ref{eq:get_queries}) --~(\ref{eq:sm_attn})) as our XQA module. 
Besides, our approach can be extended to the case of more than 2 individuals (see details in \Cref{subsec:implement} and Appendix {\color{red}B.1}).
% \cref{app_subsec:implement}). 
% If subscript $h$ presents head $h$, then the multi-head XQA becomes:
% \begin{align}
%     O^l_h & = \text{XQA}(E^l_h) \label{eq:bdattn_multi} \\
%     O^l & = \text{FC}(\text{Concat}[O^l_h])
% \end{align}
% \vskip -0.15in
% \begin{equation} \label{eq:bdattn_multi}
%     \bm{O}^l_h = \text{XQA}(\bm{E}^l_h) \hspace{3mm}\text{and}\hspace{3mm} 
%     \bm{O}^l = \text{FC}(\text{Concat}[\bm{O}^l_h]).
% \end{equation}

\subsection{Proxy}~\label{subsec:proxy}
% In addition to the fencing game, in a ball game, the motion of the ball highly influences the trajectories of the involved persons. 
% In the above two scenarios, the swords and the ball typically act as a motion intermediary or proxy, continuously interacting with the involved persons and highly influencing their trajectories. 
% To better model the scenario, we further introduce a concept of \textit{proxy} to  
% build an unit to bridge the involved persons, 
% since we assume there exists a \textit{proxy} unit for extreme actions and this concept is motivated by the following. 
% For instance, in a fencing game, the \textit{proxy} is typically the swords that closely affect the two opponents. 
% We consider the games without a \textit{proxy} as a special case (e.g., the boxing game) and can assume a virtual one for this case.
To subtly recalibrate the spatial dependencies of the body joints, we further introduce a concept of \textit{proxy} in the scenario of extreme actions which includes spatial semantic information, and integrate it into our XQA module to bridge the involved persons.
% Motivated by this, we suppose there exists a \textit{proxy} unit in the scenario of extreme actions, and integrate it into our XQA module to bridge the involved persons.
% For the case without a \textit{proxy} (e.g., dancing), we consider it as a special case but assume a virtual \textit{proxy} for it, which can be viewed as a common feature map learning the informative features from the involved persons. 
The proposed \textit{proxy} unit cooperates with the XQA module and is expected to act as a motion intermediary in the interaction modeling, subtly controlling the bidirectional information flows.
The detailed operations to construct a \textit{proxy} are given in the following. 
% To construct a \textit{proxy} unit, 
Instead of setting \textit{proxy} learnable explicitly, we let it learn and extract the poses' information from the involved persons in an implicit way. 
Specifically, $M$ trainable template vectors $t_i\in \mathbb{R}^{D} (i=1,\dots, M)$ are first introduced, where $M$ is set very small to control the number of parameters. 
These learnable vectors constitute the template matrix, denoted as $\bm{T}=[t_1, \dots, t_M] \in \mathbb{R}^{M \times D}$, which are used to transform the aggregated spatial features. 
Then we suppose that the involved persons' information influences the templates, that is, the mapping of templates is learnt from data. Consequently, 
\vskip -0.2in
\begin{equation}\label{eq:temp_w}
    \bm{W}_t = \text{FC}(\text{Concat}[\bm{E}^l, \bm{E}^f]), 
\end{equation}
where $\bm{E}^l$ and $\bm{E}^f$ are the inputs of the XQA module, concatenated by channels. 
$\bm{W}_t \in \mathbb{R}^{T \times M}$ denotes the global spatial features that aggregate the involved persons.
Then the proxy is built by transforming the aggregated global spatial information into model dimension and being formulated into spatial dependencies recalibration matrix: 
\vskip -0.05in
\begin{equation}\label{eq:proxy}
    \bm{P} = \bm{T}^{\mathsf{T}} (\bm{W}_t)^{\mathsf{T}} \bm{W}_t  \bm{T} , 
\end{equation}
% \vskip -0.05in
\noindent{where $\bm{P} \in \mathbb{R}^{D \times D}$ is the symmetric \textit{proxy} matrix, including the spatial dependencies of the body joints. } 

Finally, \textit{proxy} is implemented on the cross-query attention since it controls the bidirectional information flows and recalibrates the spatial dependencies:
\vskip -0.1in
\begin{equation} \label{eq:proxy_attn}
    \bm{A} = \bm{Q}^l \cdot \bm{P} \cdot [\bm{Q}^f]^{\mathsf{T}}, 
\end{equation}
% \vskip -0.1in
where \textit{proxy}, represented as $\bm{P}$, subtly bridges the two queries. 
It is worth noting that the future \textit{proxy} in the decoder is built by the future templates.  
Concretely, the future templates used in the decoder $\bm{T}_{de}$ can be predicted by the history one used in the encoder $\bm{T}_{en}$: 
\vskip -0.1in 
\begin{equation} \label{eq:future_temp}
    \bm{T}_{de} = \text{Attention} (\bm{T}_q, \bm{T}_{en}, \bm{T}_{en}), 
\end{equation}
where $\text{Attention} (\bm{Q}, \bm{K}, \bm{V})$ denotes the attention operation proposed by Vaswani \textit{et al.}~\cite{vaswani2017attention} with $\bm{Q}, \bm{K}, \bm{V}$ serving as query, key and value matrices, 
and $\bm{T}_q \in\mathbb{R}^{M \times D}$ is the trainable query matrix for future prediction. 
% which serves as the query templates to predict future templates. 

\subsection{Network Architecture}~\label{subsec:architecture}
\begin{figure*}[ht]
    \vskip -0.1in
    \begin{center}
        \centerline{\includegraphics[width=0.8\textwidth]{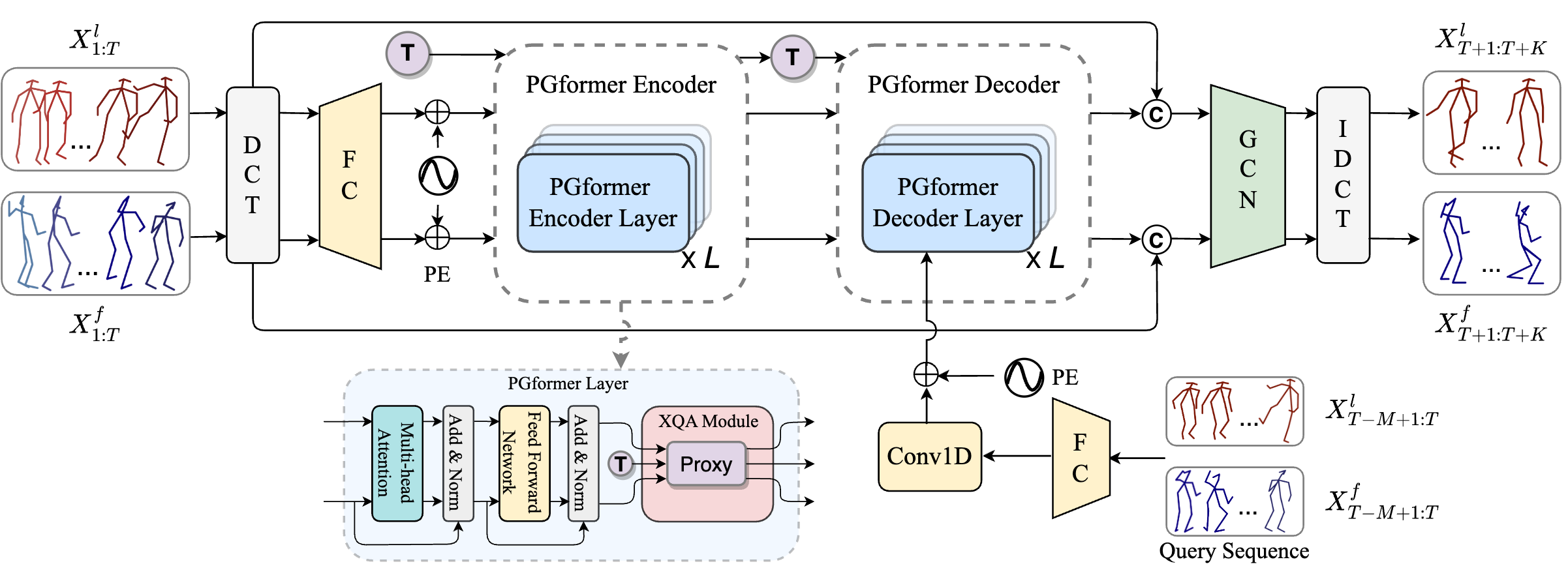}} % \linewidth
        % \vskip -0.05in
        \caption{Overview of our PGformer's architecture for multi-person highly interactive extreme motion prediction. $\oplus$ and \textcircled{c} represent broadcast element-wise addition and concatenation respectively, and PE means positional encoding. \textbf{T} denotes the template matrix used to construct \textit{proxy} in the encoder layer, and the \textit{proxy} in the decoder layer is built by the predicted future templates. The left bottom is a schematic diagram of a PGformer layer, including a standard Transformer layer (MHA + FFN) and a subsequent XQA module with \textit{proxy}. 
        }
        \label{fig:framework}
    \end{center}
    \vskip -0.4in
\end{figure*}

The overall network architecture of our PGformer is shown
in \Cref{fig:framework}. 
Our PGformer comprises three main components: a pose encoding that embeds the input 3D poses into model dimension, a non-autoregressive Transformer with cross-dependencies learning, and a pose decoding that outputs a sequence of 3D pose vectors. 
Similar to the vanilla Transformer~\cite{vaswani2017attention}, our
PGformer's encoder and decoder layers are composed of a multi-head attention mechanism (MHA), a feed-forward network (FFN), and a subsequent XQA module with \textit{proxy}, as shown in the left bottom of \Cref{fig:framework}.
Following~\cite{Martinez_potr_ICCV2021}, the decoder works in a non-autoregressive fashion to avoid error accumulation and reduce computational cost.
As the Transformer learns the temporal dependencies, the model shall identify spatial dependencies between the different body parts in the pose encoding and decoding process. 
Our approach is trained in a classifier-free and discriminator-free  fashion.
More specifically, our proposed network architecture works as follows.

% In the XQA module, \textit{proxy} transfers the effective motion information bilaterally like a bridge.

% \paragraph{PGformer Encoder.} 
\noindent\textbf{PGformer Encoder.~~}
We first apply Discrete Cosine Transform (DCT)~\cite{ahmed1974discrete,mao2019learning} to encode the input poses $\bm{X}_{1:T}$ with temporal smoothness in frequency domain. 
A fully connected (FC) layer is then used as a pose encoding network to transform the inputs with the dimension of $3J$ into the embeddings (denoted by $\bm{E}_{1:T}$) with the model dimension of $D$.
The encoder takes the sequence of pose embeddings added with positional embeddings (PE) as the inputs, which is composed of $L$ layers, each with a Transformer layer (an MHA and an FFN layers) and a subsequent XQA module. 

% \paragraph{PGformer Decoder.} 
\noindent\textbf{PGformer Decoder.~~} 
The PGformer decoder takes the encoder outputs (including $\bm{T}_{en}$) as well as a query sequence $\bm{Q}_{1:K}=[q_1, \dots, q_K]$ as inputs. 
It generates the output embeddings through a stack of $L$ PGformer layers, which is the same as in the encoder except for using a different query in the decoder.
We adopt the strategy in~\cite{guo2021multi} and let the query $q$ learn from the last $M$ frames of the input sequence $\bm{X}_{T-M+1:T}$.
Like non-autoregressive Transformer in~\cite{Martinez_potr_ICCV2021}, we use a simple approach to fill $\bm{Q}_{1:K}$ using copied entries from $q$. 
More precisely, an FC layer and a Conv1D layer are applied to transform the dimension of $\bm{X}_{T-M+1:T}$ and squeeze the sequence of length M into one vector $q \in \mathbb{R}^D$.
Then each entry $q_t$ in $\bm{Q}_{1:K}$ is a copy of $q$.
% Future predictions are generated in parallel by the network from the decoder outputs.
Lastly, we concatenate the last observation $x_T$ to the decoder outputs and apply a graph convolutional network (GCN) with a residual connection as pose decoding, which treats each joint as a node in a graph to densely learn the spatial relationships between the body joints and transform the outputs of model dimension $D$ back to the original dimension $3J$.
To obtain the final predicted poses, the Inverse Discrete Cosine Transform (IDCT) is employed on the outputs of the GCN. 

% \paragraph{Center of Gravity.}
\noindent\textbf{Gravity Loss.~~} 
In view of forecasting human motion in extreme actions, we propose a gravity loss for each person to ensure the center of gravity is kept within a plausible altitude range and avoid it dramatically vary in contiguous frames. 
In out experiment, we find that the gravity loss can improve the long-term prediction and control the variances, making the model more stable. 

Specifically, we introduce a learnable vector $w \in \mathbb{R}^{J \times 3}$ with length of $J$ and dimension of 3 (the 3D coordinates) for each person to learn the weights of each joint. 
$Softmax$ function is applied to make the summation of $J$ weights for each axis equal to 1.
Then the center of gravity at time step $t$, represented as $g_t \in \mathbb{R}^3$, is computed by: 
\vskip -0.15in
\begin{equation}\label{eq:gravity}
    g_t = \sum_{j=1}^{J} w_j \otimes \hat{x}_{t, j}, 
\end{equation}
\vskip -0.05in
\noindent{where $\otimes$ denotes broadcast element-wise multiplication, and $\hat{x}_{t,j} \in \mathbb{R}^{3}$ is the predicted pose of joint $j$ at time $t$. }
We use offset $\Delta g_t = g_{t+1} - g_{t}$ between two time steps to represent the variation. 
Then we take the summation of $\Delta g_t$ as the gravity loss in the total loss function, which is represented by $\mathcal{L}_{g^l}=\sum_{t=T}^{T+K-1} \Delta g_t^l$ for the leader, and $\mathcal{L}_{g^f}$ is the corresponding gravity loss for the follower.

%-------------------------------------------------------------------------

\section{Experiments}\label{sec:exp}
In this section, we first experimentally evaluate our approach against other state-of-the-art methods on three benchmarks, including ExPI, CMU-Mocap and MuPoTS-3D datasets.
In addition, we evaluate the model qualitatively and conduct ablation studies on ExPI. 
All models are implemented by PyTorch toolkit on a single V100 GPU.
More dataset descriptions, implementation details, experimental results, visualizations and ablation studies are provided in Appendix.

\subsection{Datasets}\label{subsec:dataset}
Most of our experiments are based on the challenging ExPI dataset containing highly interactive extreme motions. 
Besides, our approach is also validated on the CMU-Mocap and MuPoTS-3D datasets with weak interactions to inspect the transferability and the generalization ability.

% \paragraph{The Challenging ExPI Dataset.} 
\noindent\textbf{The Challenging ExPI Dataset.} 
Different from the other multi-person motion datasets, the Extreme Pose Interaction (ExPI) dataset is a special dataset of professional dancers performing Lindy-hop dancing actions, where the two dancers are called leader and follower. 
% where the human motion sequences are organized in pairs showing 16 different actions with extreme pose interactions. 
The ExPI dataset contains 2 couples of dancers performing 16 extreme actions, and it provides 115 sequences in total, in which each sequence contains 30K frames and 60K instances with annotated 3D body poses and shapes.
Each person is recorded at 25 FPS with 3D position of 18 joints. 
% The train/ test split in this experiment adopts the official settings~\cite{guo2021multi}.
Following the settings in~\cite{guo2021multi}, we conduct experiments on the data splits: common action split and unseen action split (see Appendix {\color{red}A.1} for details). 
% see \cref{app_sec:dataset}

% We present the Extreme Pose Interaction (ExPI) Dataset, a new person interaction dataset of Lindy Hop aerial steps [1].  To perform these actions, the two dancers perform different movements that require a high level of synchronization. These actions are composed of extreme poses and require strict and close cooperation between the two persons, which is highly suitable for the study of human interactions.
% Our dataset contains 2 couples of dancers performing 16 extreme actions,  obtaining 115 sequences with 30k frames for each viewpoint and 60k instances with annotated 3D body poses and shapes.

\begin{table*}[t]
    % \vskip -0.1in
    \setlength\tabcolsep{2.0pt}
    \linespread{1.3}
    \caption{Results of \textbf{JME} (MPJPE) on the common action split with the two evaluation metrics (in \textit{mm}). Lower values mean better performances. The best and second best performances are respectively marked in \textbf{bold} and \underline{underlined}. }
    % I/DCT are adopted into other models for a fair comparison.
    \vskip -0.1in
    \label{tab:expi_tab1}
    \begin{center}
    % \begin{small}
    % \footnotesize
    \scriptsize
    \begin{tabular}{l|cccc|cccc|cccc|cccc|cccc|cccc|cccc|cccc}
        % \toprule
        \hline
        % \multirow{8}{*}{\rotatebox{90}{JME}} 
        Action & \multicolumn{4}{c|}{A1 A-frame} & \multicolumn{4}{c|}{A2 Around the back} & \multicolumn{4}{c|}{A3 Coochie} & \multicolumn{4}{c|}{A4 Frog classic} & \multicolumn{4}{c|}{A5 Noser} & \multicolumn{4}{c|}{A6 Toss Out} & \multicolumn{4}{c|}{A7 Cartwheel} & \multicolumn{4}{c}{AVG} \\
        \hline 
        Time (sec) & 0.2 & 0.4 & 0.6 & 1.0 & 0.2 & 0.4 & 0.6 & 1.0 & 0.2 & 0.4 & 0.6 & 1.0 & 0.2 & 0.4 & 0.6 & 1.0 & 0.2 & 0.4 & 0.6 & 1.0 & 0.2 & 0.4 & 0.6 & 1.0 & 0.2 & 0.4 & 0.6 & 1.0 & 0.2 & 0.4 & 0.6 & 1.0 \\
        \hline
        % \multirow{6}{*}{\rotatebox{90}{JME}} 
        Res-RNN~\cite{julieta2017motion} & 83 & 141 & 182 & 236 & 127 & 224 & 305 & 433 & 99 & 177 & 239 & 350 & 74 & 135 & 182 & 250 & 87 & 152 & 201 & 271 & 93 & 166 & 225 & 321 & 104 & 189 & 269 & 414 & 95 & 169 & 229 & 325 \\
        LTD~\cite{mao2019learning} & 70 & 125 & 157 & 189 & 131 & 242 & 321 & 426 & 102 & 194 & 260 & 357 & 62 & 117 & 155 & 197 & 72 & 131 & 173 & 231 & 81 & 151 & 200 & 280 & 112 & 223 & 315 & 442 & 90 & 169 & 226 & 303 \\
        HRI~\cite{mao2020history} & 52 & 103 & 139 & 188 & 96 & 186 & 256 & 349 & 57 & 118 & 167 & 240 & 45 & 93 & 131 & 180 & 51 & 105 & 149 & 214 & 61 & 125 & 176 & 252 & 71 & 150 & 222 & 333 & 62 & 126 & 177 & 251 \\
        MSR~\cite{dang2021msr} & 56 & 100 & \underline{132} & \underline{175} & 102 & 187 & 256 & 365 & 65 & 120 & 166 & 244 & 50 & 95 & 127 & 172 & 54 & 100 & 138 & 202 & 70 & 132 & 182 & 258 & 82 & 154 & 218 & 321 & 69 & 127 & 174 & 248 \\
        SPGSN~\cite{li2022spgsn} & 68 & 132 & 173 & 245 & 114 & 215 & 272 & 376 & 83 & 155 & 211 & 299 & 63	& 118 & 159 & 209 & 66 & 134 & 182 & 245 & 75 & 148  & 205 & 297 & 95 & 185 & 263 & 388 & 81 & 155 & 209 & 294 \\
        MRT~\cite{wang2021multiperson} & 61	& 115 & 143 & 170 & 112	& 202 & 276 & 389 & 83 & 158 & 216 & 324 & 54 & 101 & 136 & 179 & 64 & 121 & 160 & 214 & 71	& 139 & 185 & 261 & 83 & 192 & 280 & 402 & 75 & 147 & 199 & 277 \\
        TBIFormer~\cite{peng2023trajectory} & 50 & 100 & 136 & 184 & 94 & 184 & 255 & 346 & 55 & 116 & 163 & 235 & 47 & 95 & 133 & 184 & 50 & 106 & 151 & 217 & 59 & 122 & 172 & 247 & 69 & 147 & 218 & 327 & 61 & 124 & 175 & 249 \\
        BP~\cite{rahman2023best} & 63 & 109 & 140 & 184 & 109 & 193 & 258 & 360 & 69 & 126 & 172 & 247 & 53 & 96 & 129 & 178 & 62 & 111 & 151 & 211 & 73 & 132 & 182 & 261 & 87 & 169 & 238 & 353 & 74 & 134 & 181 & 256 \\
        XIA~\cite{guo2021multi} & \underline{49} & \underline{98} & 140 & 192 & \textbf{84} & \underline{166} & \underline{234} & \underline{346} & \underline{51} & \underline{105} & \underline{154} & \underline{234} & \underline{41} & \underline{84} & \textbf{120} & \textbf{161} & \textbf{43} & \underline{90} & \underline{132} & \underline{197} & \underline{55} & \underline{113} & \underline{163} & \underline{242} & \underline{62} & \underline{130} & \underline{192} & \underline{291} & \underline{55} & \underline{112} & \underline{162} & \underline{238} \\
        Ours & \textbf{46} & \textbf{93} & \textbf{129} & \textbf{173} & \textbf{84} & \textbf{163} & \textbf{230} & \textbf{330} & \textbf{47} & \textbf{99} & \textbf{146} & \textbf{230} & \textbf{39} & \textbf{83} & \textbf{120} & \textbf{161} & \textbf{43} & \textbf{89} & \textbf{130} & \textbf{195} & \textbf{53} & \textbf{107} & \textbf{154} & \textbf{231} & \textbf{59} & \textbf{125} &\textbf{188} & \textbf{286} & \textbf{53} & \textbf{108} & \textbf{156} & \textbf{230} \\
        \hline
    \end{tabular}
    % \end{small}
    \end{center}
    \vskip -0.2in
\end{table*}

\noindent\textbf{CMU-Mocap and MuPoTS-3D with Weak Interactions.}
The Carnegie Mellon University Motion Capture Database (CMU-Mocap)~\cite{cmumocap} provides motion capture recordings from 140 subjects in scenarios of $1 \sim 2$ persons, performing various activities. 
The Multi-person Pose estimation Test Set in 3D (MuPoTS-3D)~\cite{mupots3d} provides 8,000 annotated frames of poses from 20 real-world scenes, and each sample contains 2 or 3 persons. 
Different from the ExPI dataset, the interactions between the persons in these two datasets are typically weak (e.g., hand-shaking, standing and chatting). 
% The experimental settings for these two datasets follow~\cite{wang2021multiperson}.
We use the derived CMU-Mocap data (3 persons), MuPoTS-3D data (2--3 persons) and Mix1 (6 persons, blended by CMU-Mocap and MuPoTS-3D) following~\cite{peng2023trajectory}.

\subsection{Evaluation Metrics}
% To better evaluate the model performances, 
We adopt the Joint Mean Error (JME) and Aligned Mean Error (AME) in~\cite{guo2021multi} as our evaluation metrics. 

% \paragraph{Joint Mean Error (JME).} 
\noindent\textbf{Joint Mean Error (JME).~~~} 
\textit{Joint Mean per joint position Error}, dubbed as JME in short, measures the average L2-norm of different persons in the same coordinate by: 
% Formally, JME is computed by: 
% The JME indistinguishably measures the poses of different persons by:
\vskip -0.1in
\begin{equation}\label{eq:jme}
    \text{JME}(\bm{P}, \bm{G}) = \text{MPJPE} (\bm{P}, \bm{G}),
    \footnote{Mean Per Joint Position Error (MPJPE) calculates the average L2-norm across different
    joints between the prediction and ground-truth, which is a widely used metric for evaluating 3D pose errors. }
    % It is computed by $\text{MPJPE}(\bm{P}, \bm{G}) = \frac{1}{KJ} \sum_{k=1}^{K} \sum_{j=1}^{J} ||\bm{P}_{k,j} - \bm{G}_{k,j}||^{2}$, where $\bm{P}_{k,j} \in \mathbb{R}^{3}$ denotes the predicted joint position of the joint $j$ in the frame $k$, and $\bm{G}_{k,j} \in \mathbb{R}^{3}$ is the corresponding ground truth.}
\end{equation}
% \vskip -0.1in \noindent
where $\bm{P}$ and $\bm{G}$ are the normalized prediction and ground truth. 
As our task aims at predicting not only the the distinct poses but also the relative position of the two person, $\bm{P}$ and $\bm{G}$ are normalized by the same person (e.g., the leader) to keep the information of their related positions. 
This considers the two interacted persons jointly as a whole and measures both the errors of poses and their relative positions.

% \paragraph{Aligned Mean Error (AME).} 
\noindent\textbf{Aligned Mean Error (AME).~~~} 
\textit{Aligned Mean per joint position Error} (or aligned MPJPE, AME for short) normalizes the data by removing the global movement of the poses based on a selected root joint (Procrustes analysis~\cite{gower1975generalized}) before computing MPJPE. 
Formally, AME is computed by:
\vskip -0.1in
\begin{equation}
    \text{AME}(\hat{\bm{P}}, \hat{\bm{G}}) = \text{MPJPE}(T_A(\hat{\bm{P}}, \hat{\bm{G}}), \hat{\bm{G}}),
\end{equation}
% \vskip -0.1in \noindent
where $\hat{\bm{P}}$ and $\hat{\bm{G}}$ are the independently normalized poses to erase the errors of the relative positions between the two persons. 
$T_A$ is a rigid alignment function between the estimated pose and ground truth proposed in~\cite{gower1975generalized}, further mitigating the impacts of the joints that are used to determine the coordinate (hips and back).

\begin{table*}[ht]
\setlength\tabcolsep{1.8pt}
\linespread{1.1}
\vskip -0.2in
\caption{Action-wise JME results (in \textit{mm}) on the unseen action split, where the testing actions do not appear in the training.}
% The best and second best performances are respectively marked in \textbf{bold} and \underline{underlined}.}
% \vskip -0.2in
\label{tab:expi_tab3}
\begin{center}
\footnotesize
% \scriptsize
\begin{tabular}{l|ccc|ccc|ccc|ccc|ccc|ccc|ccc|ccc|ccc|ccc}
    \hline
    % \toprule
    Action & \multicolumn{3}{c|}{A8} & \multicolumn{3}{c|}{A9} & \multicolumn{3}{c|}{A10} & \multicolumn{3}{c|}{A11} & \multicolumn{3}{c|}{A12} & \multicolumn{3}{c|}{A13} & \multicolumn{3}{c|}{A14} & \multicolumn{3}{c|}{A15} & \multicolumn{3}{c|}{A16} & \multicolumn{3}{c}{AVG} \\
    \hline
    Time (sec) & 0.2 & 0.6 & 1.0 & 0.2 & 0.6 & 1.0 & 0.2 & 0.6 & 1.0 & 0.2 & 0.6 & 1.0 & 0.2 & 0.6 & 1.0 & 0.2 & 0.6 & 1.0 & 0.2 & 0.6 & 1.0 & 0.2 & 0.6 & 1.0 & 0.2 & 0.6 & 1.0 & 0.2 & 0.6 & 1.0 \\
    \hline
    % JME
    HRI~\cite{mao2020history} & 57 & 188 & 278 & 49 & 105 & 137 & 55 & 153 & 227 & 80 & 217 & 280 & 68 & \textbf{183} & \textbf{256} & 45 & 148 & 246 & 93 & 283 & 438 & 61 & 165 & 240 & 49 & 132 & 197 & 62 & 175 & 255 \\
    MSR~\cite{dang2021msr} & \textbf{54} & 177 & 269 & 53 & 123 & 168 & \textbf{53} & \textbf{150} &  \textbf{218} & 79 & 221 & 311 & 70 & 190 & 274 & 43 & 148 & 250 & 95 & 278 & 414 & 61 & 174 & 263 & 50 & 142 & 217 & 62 & 177 & 264 \\
    XIA~\cite{guo2021multi} & 56 & 181 & 274 & 49 & 108 & 139 & 55 & 153 & 222 & 79 & 213 & 282 & 68 & 184 & 260 & 44 & 147 & 243 & 93 & 272 & 410 & 61 & \textbf{160} & \textbf{230} & \textbf{48} & \textbf{130} & 193 & 61 & 172 & 250 \\
    % SPGSN~\cite{li2022spgsn} \\
    MRT~\cite{wang2021multiperson} & 57 & 183 & 274 & 50 & 108 & 143 & 54 & 152 & 221 & 79 & 216 & 288 & 69 & 185 & 267 & 45 & 149 & 250 & 94 & 277 & 412 & 61 & 168 & 246 &49 & 138 & 210 & 62 & 175 & 257 \\
    TBIFormer~\cite{peng2023trajectory} & 56 & 179 & 270 & \textbf{48} & 106 & 138 & 54 & 153 & 223 & \textbf{78} & 210 & 277 & 70 & 188 & 271 & 44 & 148 & 248 & 94 & 279 & 416 & \textbf{60} & 164 &	234 & 49 & 136 & 205 & 61 & 174 & 254 \\
    BP~\cite{rahman2023best} & 64 & 190 & 275 & 60& 127 & 165 & 66 & 162 & 233 & 90 & 232 & 305 & 84 & 212 & 291 & 59 & 156 & 239 & 115 & 293 & 415 & 76 & 189 & 266 & 60 & 151 & 215 & 75 & 190 & 267 \\
    Ours & \textbf{54} & \textbf{176} & \textbf{265} & \textbf{48} & \textbf{104} & \textbf{136} & \textbf{53} & 152 & 220 & \textbf{78} &  \textbf{205} & \textbf{260} & \textbf{65} & \textbf{183} & \textbf{256} & \textbf{42} & \textbf{142} & \textbf{220} & \textbf{90} & \textbf{266} & \textbf{408} & \textbf{60} & 162 & 232 & \textbf{48} & \textbf{130} & \textbf{191} & \textbf{60} & \textbf{169} & \textbf{243} \\
    \hline
    % AME
    % & HRI~\cite{mao2020history} & 35 & 106 & \underline{151} & \underline{28} & 75 & \underline{97} & 31 & 88 & 124 & 45 & 121 & 160 & 38 & 106 & 149 & 25 & 80 & 128 & 49 & 147 & 211 & \underline{31} & 82 & 106 & 28 & \underline{70} & \textbf{94} & \underline{34} & 97 & 135 \\
    % \rowcolor{mygray} \cellcolor{white} & MSR~\cite{dang2021msr} & \textbf{34} & \textbf{103} & \textbf{150} & 31 & 85 & 116 & 31 & \textbf{87} & \textbf{114} & \underline{44} & 119 & 163 & 37 & 105 & 151 & \underline{24} & 80 & 132 & 50 & 143 & 202 & 32 & 86 & 116 & 29 & 75 & 106 & 35 & 98 & 139 \\
    % & XIA~\cite{guo2021multi} & \textbf{34} & \underline{104} & 153 & \underline{28} & \underline{73} & \textbf{96} & 31 & \textbf{87} & 117 & \underline{44} & \underline{118} & \underline{158} & \textbf{36} & \underline{102} & \underline{142} & 25 & 80 & \underline{125} & 48 & 141 & 204 & \textbf{30} & \textbf{78} & \textbf{103} & \textbf{27} & \underline{70} & \underline{95} & \underline{34} & \underline{95} & \underline{132} \\
    % \rowcolor{mygray} \cellcolor{white} \multirow{-4}{*}{\rotatebox{90}{AME}} & Ours & \textbf{34} & \textbf{103} & \textbf{150} & \textbf{27} & \textbf{72} & \underline{97} & 31 & 88 & \underline{116} & \textbf{43} & \textbf{117} & \textbf{152} & \textbf{36} & \textbf{100} & \textbf{140} & \textbf{23} & \textbf{73} & \textbf{109} & \textbf{45} & \textbf{135} & \textbf{196} & \underline{31} & \underline{81} & \underline{104} & \textbf{27} & \textbf{69} & \textbf{94} & \textbf{33} & \textbf{94} & \textbf{129} \\
    % \hline
    % \bottomrule
\end{tabular}
\end{center}
\vskip -0.15in
\end{table*}

\begin{table*}[t]
    \caption{Results of MPJPE on CMU-Mocap and MuPoTS-3D.}
    % \vskip -0.2in
    \label{tab:cmu_mup}
    \begin{center}
    \footnotesize
    % \small
    \begin{tabular}{l|ccc|ccc|ccc}
        % \toprule
        \hline
        & \multicolumn{3}{c|}{CMU-Mocap (3 persons)} & \multicolumn{3}{c|}{MuPoTS-3D (2--3 persons)} & \multicolumn{3}{c}{Mix1 (6 persons)}  \\
        % \hline
        \cline{2-10}
        Method & 0.2s & 0.6s & 1.0s & 0.2s & 0.6s & 1.0s & 0.2s & 0.6s & 1.0s \\
        \hline
        % \midrule
        HRI~\cite{mao2020history} & 49 & 130 & 207 & 81 & 211 & 323 & 51 & 141 & 233 \\
        MSR~\cite{dang2021msr} & 53 & 146 & 231 & 79 & 222 & 374 & 49 & 132 & 220 \\
        MRT~\cite{wang2021multiperson} & 36 & 115 & 192 & 78 & 225 & 349 & 37 & 122 & 212 \\
        TBIFormer~\cite{peng2023trajectory} & \textbf{30} & 109 & 182 & \textbf{66} & 200 & 319 & \textbf{34} & 121 & 209 \\
        Ours & 31 & \textbf{108} & \textbf{179} & 68 & \textbf{197} &  \textbf{315} & 35 & \textbf{120} & \textbf{207} \\
        \hline
        % \bottomrule
    \end{tabular}
    \end{center}
    \vskip -0.2in
\end{table*}

\begin{figure}[t]
    \begin{center}
        \centerline{\includegraphics[width=0.8\linewidth]{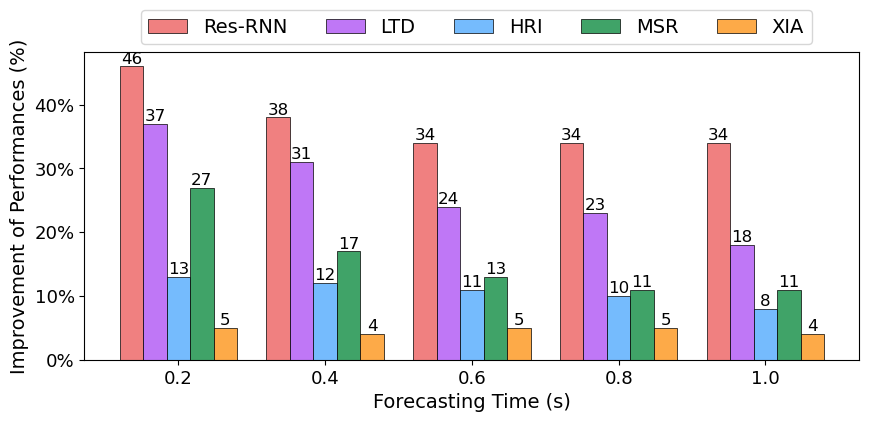}} % \linewidth
        % \vskip -0.1in
        \caption{Percentages of improvement of our PGformer compared with other methods at different forecasting time, on the common action split, which are measured by taking the average of the percentages of improvement of average JME and AME error.}
        \label{fig:barchart}
    \end{center}
    \vskip -0.4in
\end{figure}

\subsection{Implementation Details} 
\label{subsec:implement}
Our proposed architecture has $L$ ($L=4$) PGformer layers in the encoder and decoder with a dimension of $D$ ($D=128$), and $M$ ($M=3$) learnable template vectors are used to construct the \textit{proxy}. 
For training our PGformer on ExPI, we follow the same implementation settings as in~\cite{guo2021multi}. 
% Specifically, we predict future motion for 1 second in a recursive manner based on the observed motion of 50 frames. 
% The network is trained by the Adam optimizer with an initial learning rate of 0.005, which is decayed by a rate of $0.1^{1/E}$ ($E$ is the total number of epochs) every epoch. 
A gravity loss is added to control the variation of center of gravity for each person on the original loss function: $\mathcal{L} = \mathcal{L}_f + 10^{-\epsilon} \mathcal{L}_l + \lambda_l \mathcal{L}_{g^l} + \lambda_f \mathcal{L}_{g^f}$, where $\mathcal{L}_f$ and $\mathcal{L}_l$ are the average MPJPE loss for the leader and follower respectively,  and $\epsilon$ denotes epoch index. 
We set the weights of gravity losses $\lambda_l$ and $\lambda_f$ as 0.01 and 0.0001, respectively.

For CMU-Mocap and MuPoTS-3D, the implementation settings exactly follow~\cite{wang2021multiperson} except that the inputs are absolute coordinates ($x_t$) instead of motions ($\Delta x_t$). 
% Under their protocols, 
The model is trained on a synthesized dataset mixing sampled motions from CMU-Mocap to create 3-person scenes and evaluated on both CMU-Mocap and MuPoTS-3D datasets. 
In this case, each person is denoted as $\bm{E}^l$, and other persons are concatenated by time as $\bm{E}^f$ (e.g., 3 persons mean 3 pairs of $\bm{E}^l$ and $\bm{E}^f$). 
This implementation can be adaptive to any number of persons regardless of parameters.
The attention score map $\bm{A} \in \mathbb{R}^{T \times ((n-1) \times T)}$, where $n$ is the number of persons, could still be shared by $\bm{Q}^l$ and $\bm{Q}^f$, but it only be used to obtain $\bm{O}^l$ for simplicity.
The entire process is conducted in an iterative manner over $n$ with the shared parameters (see Appendix {\color{red}B.1} for more details). 
% The gravity loss is not applied since the motions in the two datasets are moderate.

\subsection{Quantitative Results}

\begin{figure}[t]
    \begin{center}
        \centerline{\includegraphics[width=1.0\linewidth]{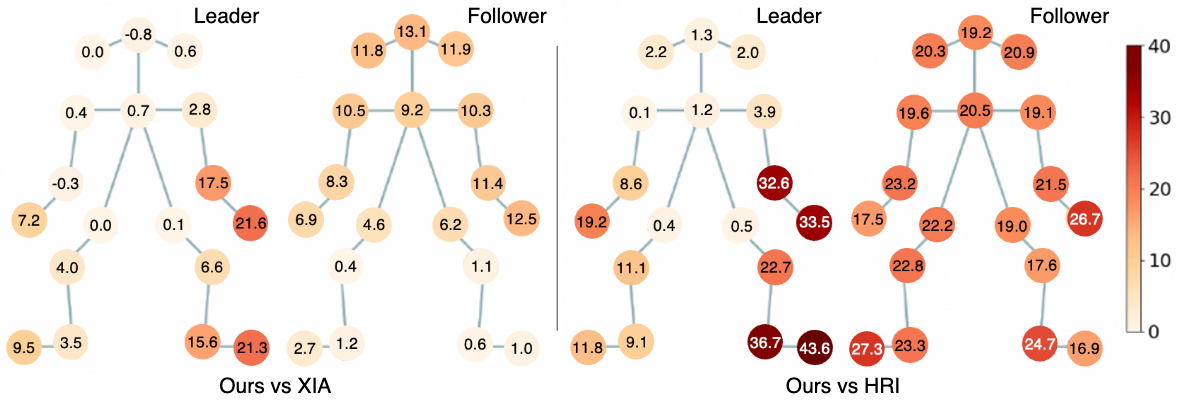}} % \linewidth
        % \vskip -0.1in
        \caption{Average performance gains over XIA and HRI of joint-wise JME on ExPI. Darker color means larger performance gains.}
        \label{fig:jointgain}
    \end{center}
    \vskip -0.4in
\end{figure}

\subsubsection{Results on ExPI}
We compare ours with Res-RNN~\cite{julieta2017motion}, LTD~\cite{mao2019learning}, HRI~\cite{mao2020history}, MSR-GCN (MSR)~\cite{dang2021msr}, and XIA-GCN (XIA)~\cite{guo2021multi}, 
of which the results are from~\cite{guo2021multi} in terms of JME and AME in millimeter (\textit{mm}). 
We also compare ours with SPGSN~\cite{li2022spgsn}, MRT~\cite{wang2021multiperson}, BestPractices (BP)~\cite{rahman2023best} and TBIFormer~\cite{peng2023trajectory}, whose results are produced by the training protocol of~\cite{guo2021multi}.
% In the following, MSR and XIA are used to represent MSR-GCN and XIA-GCN for space limitation. 
% I/DCT are adopted into these models for a fair comparison.

% \paragraph{Common action split.}
\noindent\textbf{Common action split.~~}
The results on the common action split of ExPI are reported in \Cref{tab:expi_tab1}.
We observe that our proposed PGformer consistently outperforms other methods almost for all actions, in all metrics, in both short- (0.2--0.4 sec) and long-term (0.6--1.0 sec) predictions. 
Considering the average errors, our PGformer surpasses other state-of-the-art methods in short- and long-term predictions by 4--46\% and 4--34\% in terms of percentage of improvement on average JME and AME (\Cref{fig:barchart}).
Even though our model does not achieve the best performances on three actions in long-term forecasting, our approach still obtains the second lowest errors and has comparable performances. 
Note that, at the horizon of 1.0 sec, for the action of A5/A2, though our model achieves the second-best performance in terms of JME/AME, ours performs the best on the other metric. 
% Improvements are less in the case of periodic actions such as Walking (17\%) but larger for aperiodic actions such as Posing (40\%).
% {\color{red} add joint gain plot description, see appendix}
It is worth noting that MRT~\cite{wang2021multiperson} and TBIFormer~\cite{peng2023trajectory} are multi-person-based models proposed on weakly interacted actions, verifying that these methods cannot well model actions with extreme motions. 

We also examine the performance gains of ours over XIA and HRI for each joint in \Cref{fig:jointgain}. 
% where deeper red color means higher performance gain. 
As can be seen, our proposed method gets better results almost on all the joints, and larger performance gains are achieved for the joints of limbs. 
Since joints on the limbs usually have higher motion frequencies, the figure indicates that our PGformer can better handle high-frequency motions.
Comparing ours and XIA on the follower, larger improvements are achieved for joints on the head and shoulder. 
We reasonably conjecture that the follower has more extreme motions in Lindy-hop dancing actions (see the qualitative results for verification), and our approach can better handle extreme motions.

% \paragraph{Unseen action split.}
\noindent\textbf{Unseen action split.~~}
The results on the unseen action split are given in \Cref{tab:expi_tab3} to measure the generalization ability of models since the testing actions do not appear in the training process. 
Since the results of the unseen action split in~\cite{guo2021multi} are found to be inconsistent with the forecasting time, we reproduce the compared models and report the results in \Cref{tab:expi_tab3}. 
PGformer almost achieves the best and second-best on most of the actions across different forecasting time though the model has never `seen' the testing actions during the training process, verifying our's generalization ability.
% For the single action split, the results are worse than the corresponding ones on the common action split. 
% This is probably because training on different actions contributes to the robustness of the models, and the data of single action are insufficient to train a deep learning model.
% meaning that training on different actions helps regularise the network for this very challenging collaborative extreme motion prediction task.
% Regarding the unseen action split, 

\subsubsection{Results on CMU-Mocap and MuPoTS-3D}
% The same comparisons are conducted on the CMU-Mocap and MuPoTS-3D datasets, as shown in \Cref{tab:cmu_mup}. 
We additionally evaluate our PGformer on the derived CMU-Mocap, MuPoTS-3D and Mix1 datasets, comparing ours with several state-of-the-art methods, including HRI~\cite{mao2020history}, MSR~\cite{dang2021msr}, MRT~\cite{wang2021multiperson} and TBIFormer~\cite{peng2023trajectory}. 
It is worth noting that HRI and MSR are single-person-based methods, while MRT and TBIFormer are multi-person-based methods. 
\Cref{tab:cmu_mup} reports the results in MPJPE at 0.2, 0.6, and 1.0 seconds in the future, and the results of the compared models are from~\cite{peng2023trajectory}. 
% Using their protocols, models are trained on a synthesized dataset mixing sampled motions from CMU-Mocap to create 3-person scenes and evaluated on both CMU-Mocap and MuPoTS-3D.
% The performance gains brought by our PGformer can be consistently observed on these two datasets at different forecasting time, 
We can observe that our PGformer and SOTA model TBIFormer perform comparably at different time steps on all datasets, where ours brings more performance gains in long-term prediction. 
These observations demonstrate that our proposed method can be well generalized to weakly interacted actions and the case of more than 2 individuals.

\subsection{Qualitative Results}

\begin{figure*}[ht]
    % \vskip -0.1in
    \begin{center}
        \centerline{\includegraphics[width=0.9\textwidth]{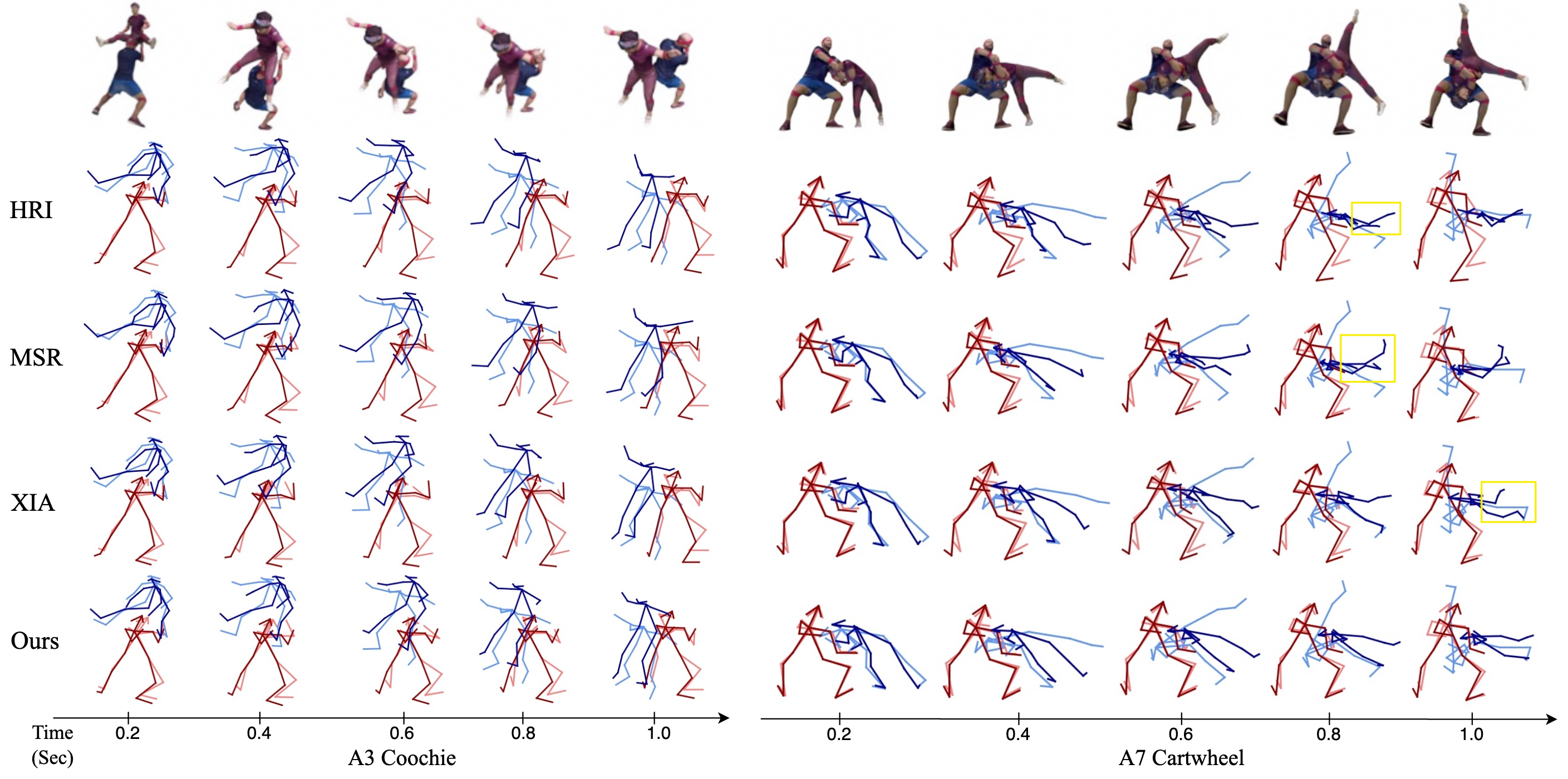}} % \linewidth
        \vskip -0.1in
        \caption{\textbf{Qualitative comparisons with other methods.} 
        \textbf{1st row:} 3D sample meshes from ExPI Dataset (just for visualization purposes). 
        \textbf{2nd-5th rows:} Motion results predicted by HRI~\cite{mao2020history}, MSR-GCN~\cite{dang2021msr}, XIA-GCN~\cite{guo2021multi}, and our PGformer. 
        Dark red/blue represents the prediction results, while light red/blue indicates the ground truths. 
        Our approach of highly interactive motion prediction achieves significantly better results than other methods. 
        More qualitative examples could be found in Appendix.}
        \label{fig:visual_a3a7}
    \end{center}
    \vskip -0.4in
\end{figure*}

% Qualitative comparisons are provided in \Cref{fig:visual_a3a7}. 
\Cref{fig:visual_a3a7} shows some qualitative results, comparing our PGformer with HRI, MSR, XIA and the ground truths, on the common action split. 
It can be seen that our predicted poses are more natural and smoother while being much closer to the ground truths than the other methods. 
Owing to effectively exploring the interactions between the collaborative persons, our PGformer performs well even on some extreme actions where other methods totally fail. 
More specifically, see the cases of action A7 in \Cref{fig:visual_a3a7}, our approach precisely forecasts the actions in long-term prediction while other methods fail to catch the motions. See more in {\color{red}B.4}. 
% More qualitative results are provided in Appendix {\color{red}B.4}.
% Our approach achieves significantly better results than other methods that independently predict the motion of each person (HRI~\cite{mao2020history} and MSR-GCN~\cite{dang2021msr}) or only study the interactions between the historical motions.

\subsection{Ablation Study}
We ablate different proposed components based on the baseline Transformer (BT) on common actions to identify their roles.
% Specifically, we compare different variants with our proposed one by ablating the main components, different layers and model dimensions. 
The results in \Cref{tab:ablation} validate the effectivenesses of our proposed \textit{proxy}, XQA module and gravity loss. 
Remarkably, the XQA module can boost the transformer without interactions notably in long-term prediction, especially on JME, demonstrating our motivation of learning the interactions between the involved persons. 
Additionally, the gravity loss can improve the long-term prediction and control the variances, making the model more stable. 
Besides, we also examine the suitableness of our suggested baseline Transformer's architecture, which has 4 PGformer layers in the encoder/decoder with D = 128 and $d_{ffn}$=1024 for model dimension and FFN, and 4 heads in MHA with $d_h$=64 for dimension of each head.
More ablation study and hyperparameter tunings are given in {\color{red}B.5}.
% \Cref{tab:ablation2}.

\begin{table}[t]
    \setlength\tabcolsep{2.8pt}
    % \vskip -0.05in
    \caption{Results of JME (MPJPE) for the compared variants. 
    The mean and standard deviation, denoted as avg and std, are computed by 5 runs. 
    BT means the baseline Transformer model. 
    $P$ and $\Delta g_t$ indicate the \textit{proxy} and gravity loss. }
    \vskip -0.2in
    \label{tab:ablation}
    \begin{center}
    \footnotesize
    % \small
    \begin{tabular}{l|cccc}
        \hline
        Time (sec) & 0.2 & 0.4 & 0.6 & 1.0 \\
        \hline
        PGformer avg ($\pm$ std) & 53 ($\pm$ 0.0) & 108 ($\pm$ 0.4) & 156 ($\pm$ 1.2) & 231 ($\pm$ 1.4) \\
        ~~~- w/o I/DCT & 57 & 113 & 161 & 234 \\
        \hline
        BT avg ($\pm$ std) & 54 ($\pm$ 0.5) & 112 ($\pm$ 0.5) & 166 ($\pm$ 1.8) & 247 ($\pm$ 2.2) \\
        ~~~+ $\Delta g_t$ avg ($\pm$ std) & 53 ($\pm$ 0.0) & 110 ($\pm$ 0.4) & 163 ($\pm$ 0.9) & 244 ($\pm$ 1.2) \\
        ~~~+ XQA & 54 & 110 & 159 & 235 \\
        ~~~+ XQA + $P$ & 53 & 108 & 157 & 232 \\
        ~~~+ XQA + $P$ + $\Delta g_t$ & 53 & 108 & 156 & 231 \\ 
        \hline
        $D$=128, $H$=4, $d_h$=32 & 54 & 110 & 160 & 238 \\
        $D$=128, $H$=8, $d_h$=32 & 53 & 109 & 157 & 233 \\
        $D$=256, $H$=8, $d_h$=32 & 53 & 109 & 159 & 237 \\
        $D$=256, $H$=8, $d_h$=64 & 53 & 109 & 160 & 239 \\
        \hline
    \end{tabular}
    \end{center}
    \vskip -0.2in
\end{table}

%------------------------------------------------------------------------
\section{Conclusions}\label{sec:conclusion}
This paper focuses on multi-person pose forecasting in a real-world scenario with highly interactive motions. 
A simple yet effective Transformer-based framework called PGformer is proposed for multi-person scenario modeling. 
Specifically, a bespoke XQA module is first proposed to learn the cross-dependencies bidirectionally between the involved persons by a shared attention score map. 
Since there typically exists a proxy continuously affecting the highly interacted persons, a concept of \textit{proxy} is introduced. 
Cooperating with the XQA module, the \textit{proxy} built by learnable templates can provide a subtle control of the bidirectional information flows from the past to the future, transferring the effective pose information bilaterally like a bridge. 
The resulting model with the above designs explores the interactions not only in learning the historical poses but in generating the unknown follow-up motions as well. 
Superior experimental results in both short- and long-term motion predictions on ExPI verify the effectiveness of our PGformer. 
We also show that our approach can be well-compatible with weakly interacted datasets.
% In the XQA module, \textit{proxy} transfers the effective motion information bilaterally like a bridge.

%------------------------------------------------------------------------
%%%%%%%%% REFERENCES
% \newpage % \clearpage works better
\clearpage
{\small
\bibliographystyle{ieee_fullname}
\bibliography{pgformer}
}

%%%%%%%%%%%%%%%%%%%%%%%%%%%%%%%%%%%%%%%%%%%%%%%%%%
%%%%%%%%%%%%%%%%%%%%%%%%%%%%%%%%%%%%%%%%%%%%%%%%%%

%%%%%%%%%%%%%%%%%%%%%%%%%%%%%%%%%%%%%%%%%%%%%%%%%%
%%%%%%%%%%%%%%%%%%%%%%%%%%%%%%%%%%%%%%%%%%%%%%%%%%
\clearpage
\appendix
\section*{Appendix}
% \title{PGformer: Supplementary Material}
% \maketitle
%%%%%%%%%%%%%%%%%%%%%%%%%%%%%%%%%%%%%%%%%%%%%%%%%%
%%%%%%%%%%%%%%%%%%%%%%%%%%%%%%%%%%%%%%%%%%%%%%%%%%
This appendix contains supplementary explanations and experiments to support our proposed proxy-bridged game Transformer (PGformer).
\Cref{app_sec:dataset} supplements the settings for the three datasets, including the descriptions of the ExPI, illustrations for the three splits, and explanations for CMU-Mocap and MuPoTS-3D settings used in our experiments. 
\Cref{app_sec:exp} provides more experiment details, results, ablation studies and visualizations. 

\section{More Information about the Dataset}
\label{app_sec:dataset}
\subsection{ExPI Settings}
As described in Section 4.1, 16 actions are recorded in the ExPI dataset, which are split into three data splits: common action split, single action split and unseen action split.
Seven of them are common actions (A1--A7), performed by both of the 2 couples. 
In our experiment, we mainly forcus on the common action split and unseen action split. 

We use superscript and subscript to denote the couple number and action split respectively, for example, the common action performed by couple 1 is denoted as $\mathcal{A}_c^1$.
The other nine actions are couple-specific and performed by only one of the couples.
The actions A8--A13 in unseen action split are performed by couple 1, denoted as $\mathcal{A}_u^1$; while the actions A14--A16 performed by couple 2 are represented as $\mathcal{A}_u^2$. 

\paragraph{Common action split.}
The common actions performed by different couples of actors are considered as training and testing data.
Then, training and testing sets contain the same actions but are performed by different persons.
In our experiment, following the setting in \cite{guo2021multi}, $\mathcal{A}_c^2$ is the training set and $\mathcal{A}_c^1$ is the testing set.

\paragraph{Single action split.}
In this split, 7 action-wise models are trained independently for each common action by treating the action from couple 2 as the training set and the same action from couple 1 as the corresponding testing set.

\paragraph{Unseen action split.}
The entire set of common actions including $\mathcal{A}_c^1$ and $\mathcal{A}_c^2$ are used as the training set for unseen action split, while the unseen actions $\{\mathcal{A}_u^1, \mathcal{A}_u^2\}$ are used as the testing set.
Since the testing actions do not appear in the training process, this unseen action split aims at measuring the generalization ability of models.

\subsection{CMU-Mocap and MuPoTS-3D Settings}
CMU-Mocap contains a large number of scenes with a single person moving and a small number of scenes with two persons interacting and moving. 
Wang \textit{et al.}~\cite{wang2021multiperson} sampled from these two parts and mix them together as their training data.
All the CMU-Mocap data were made to consist of 3 persons in each scene, and the testing set was sampled from CMU-Mocap in a similar way. 
The generalization ability of the model is evaluated by testing on the MuPoTS-3D (2 -- 3 persons) and Mix1 (6 persons) datasets with the model trained on the entire CMU-Mocap dataset.

\section{Experiments}
\label{app_sec:exp}

% \begin{figure*}[ht]
% 	\begin{center}
% 		\centerline{\includegraphics[width=0.9\textwidth]{jointgain_barchart.png}} % \linewidth
%         \vskip -0.1in
% 		\caption{\textbf{Left:} Average performance gain over XIA and HRI of joints. \textbf{Right:} Percentages of improvement of our method compared with other methods at different forecasting time, on the common action split, which are measured by taking the average of the percentages of improvement of average JME and AME error.}
% 		\label{fig:joint_bar}
% 	\end{center}
%   \vskip -0.4in
% \end{figure*}

\subsection{More Implementation Details}
\label{app_subsec:implement}
\paragraph{ExPI} 
For training our PGformer on ExPI, we follow the same implementation settings as in~\cite{guo2021multi}. 
Specifically, we predict future motion for 1 second in a recursive manner based on the observed motion of 50 frames. 
The network is trained by the Adam optimizer with an initial learning rate of 0.005, which is decayed by a rate of $0.1^{1/E}$ ($E$ is the total number of epochs) every epoch. 
Our model is trained for 40 epochs with a batch size of 32, and the average MPJPE loss is calculated for 10 predicted frames. 
And we find that XIA-GCN~\cite{guo2021multi} also has to be trained by 40 epochs to achieve the reported results. 

\paragraph{CMU-Mocap and MuPoTS-3D} 
The model predicts the future 45 frames (3 s) given 15 frames (1 s) of history as input. 
All the persons' pose sequences are forwarded in parallel to the PGformer layers to capture fine relations across themselves and other persons. 
The gravity loss is not applied to control the center of gravity since the motions in the two datasets are moderate. 

Since these two datasets consist of 2--3 persons in each scene, our XQA module should be made adaptive to them. 
Specifically, each person is denoted as $\bm{E}^l$, and other persons are concatenated by time as $\bm{E}^f$ (e.g., 3 persons mean 3 pairs of $\bm{E}^l$ and $\bm{E}^f$). 
This implementation can be adaptive to any number of persons regardless of parameters. 
The attention score map $\bm{A} \in \mathbb{R}^{T \times ((n-1) \times T)}$, where $n$ is the number of persons, could still be shared by $\bm{Q}^l$ and $\bm{Q}^f$, but it only be used to obtain $\bm{O}^l$ for simplicity ($\bm{O}^f$ is omitted).
The entire process is conducted in an iterative manner over $n$ with the shared parameters. 
Here we just provide a straightforward solution for $\geq 3$ extension, and this approach can be easily applied to the scenarios with more than 3 individuals. 
Instead of squeezing $M$ frames $\bm{X}_{T-M+1:T}$ into one vector $q$, we use the last frame $x_t$ as $q$ directly.

\subsection{More Discussions on Quantitative Results}

\paragraph{ExPI.} 
We further compare the performance gains of our PGformer with XIA-GCN~\cite{guo2021multi} and HRI~\cite{mao2020history} for each joint in Figure {\color{red}5}. 
As can be seen, our proposed method gets better results almost on all the joints, and larger performance gains are achieved for the joints of limbs. 
Since joints on the limbs usually have higher motion frequencies, the figure indicates that our PGformer can better handle high-frequency motions.
Comparing ours and XIA-GCN on the follower, larger improvements are achieved for joints on the head and shoulder. 
We reasonably conjecture that the follower has more extreme motions in Lindy-hop dancing actions (see qualitative results for verification), and our approach can better handle extreme motions.

For SPGSN, we apply it adaptively to the ExPI dataset, and decompose the body joints into upper body and lower body following the same spirit as in its experiments on Human3.6M, CMU Mocap and 3DPW datasets. 

% We conduct some additional experiments for more innovative comparisons. 
% SPGSN~\cite{li2022spgsn} is proposed for single-person motion prediction, and 
Though BP~\cite{rahman2023best} is a \textbf{contemporaneous work}, we still compare ours with BP on ExPI in \Cref{tab:more_results}. 
% and will include them in our final version. 
Since BP used different data and training settings from them used in other models (e.g., XIA, MSR and HRI), we train BP by the training setup provided by ExPI benchmark~\cite{guo2021multi} for fair comparisons. 
We also train our PGformer by the training settings provided by BP (see the results of BP trained by XIA and PGformer trained by BP). 
Besides, BP concatenates the joints of the two persons as the nodes of GCN and apply the spatial-temporal GCN, which means the number of persons should be fixed, while our PGformer can be adaptive to different numbers of persons. 
% Besides, we run the code provided by BP GitHub directly \textbf{5} times to compute the mean (avg) and standard deviation (std), and found that the stds are large. 
% The best results we obtained performed much worse than the ones reported in that paper, and the model even failed to converge for 2 times. 

\begin{table}[ht]
    \setlength\tabcolsep{2.8pt}
    \caption{Results of MPJPE for the compared models. 
    The mean and standard deviation, denoted as avg and std, are computed by 5 runs. 
    % We run the code provided by \textbf{BP official GitHub} and report the results in brackets.
    We run the code provided by \textbf{BP official GitHub} directly and report the results in brackets since it has experiments on ExPI. 
    % For SPGSN, we train the model provided by its official GitHub using the training settings in~\cite{guo2021multi}.
    }
    \vskip -0.05in
    \label{tab:more_results}
    \begin{center}
    \footnotesize
    % \small
    \begin{tabular}{l|cccc}
        % \toprule
        \hline
        Time (sec) & 0.2 & 0.4 & 0.6 & 1.0 \\
        \hline
        PGformer avg $\pm$ std & 53 $\pm$ 0.0 & 108 $\pm$ 0.4 & 156 $\pm$ 1.2 & 231 $\pm$ 1.4 \\
        PGformer (trained by BP) & 48 & 100 & 149 & 229 \\
        BP-paper (our run) & 39 (46) & 86 (97) & 129 (145) & 202 (225) \\
        % BP avg $\pm$ std (2 runs failed) & 54 $\pm$9.4 & 112 $\pm$18.2& 162 $\pm$23.8& 243 $\pm$28.5 \\
        BP (trained by XIA) & 74 & 134 & 181 & 256 \\
        \hline
    \end{tabular}
    \end{center}
    \vskip -0.2in
\end{table}

\subsection{More Comparisons on Quantitative Results}~\label{app_subsec:more_results}
\begin{figure*}[t]
	\begin{center}
		\centerline{\includegraphics[width=0.8\linewidth]{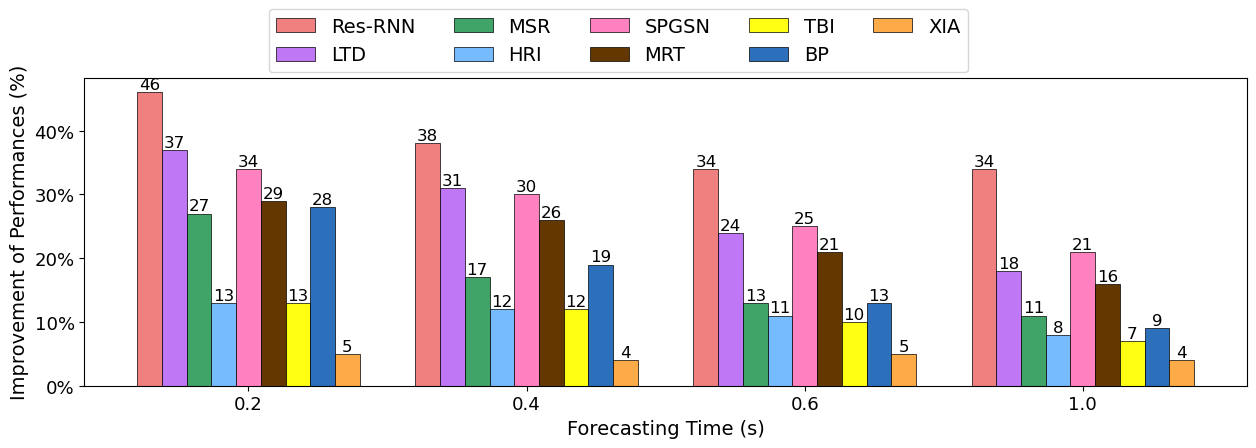}} % \linewidth
        % \vskip -0.1in
		\caption{Percentages of improvement of our PGformer compared with other methods at different forecasting time, on the common action split, which are measured by taking the average of the percentages of improvement of average JME and AME error.}
		\label{fig:barchart_v2}
	\end{center}
\end{figure*}

Due to the limited space in main paper, we remove some results of AME in Appendix \Cref{tab:expi_tab1_ame}. And we provide a more complete percentages of improvement of our PGformer compared with other methods at different forecasting time in \Cref{fig:barchart_v2}.

\begin{table*}[t]
    \vskip -0.1in
    \setlength\tabcolsep{1.8pt}
    \linespread{1.3}
    \caption{Results of \textbf{AME} on the common action split with the two evaluation metrics (in \textit{mm}). Lower values mean better performances. The best and second best performances are respectively marked in \textbf{bold} and \underline{underlined}. }
    % \vskip -0.2in
    \label{tab:expi_tab1_ame}
    \begin{center}
    % \begin{small}
    % \footnotesize
    \scriptsize
    \begin{tabular}{l|cccc|cccc|cccc|cccc|cccc|cccc|cccc|cccc}
        % \toprule
        \hline
        % \multirow{8}{*}{\rotatebox{90}{JME}} 
        Action & \multicolumn{4}{c|}{A1 A-frame} & \multicolumn{4}{c|}{A2 Around the back} & \multicolumn{4}{c|}{A3 Coochie} & \multicolumn{4}{c|}{A4 Frog classic} & \multicolumn{4}{c|}{A5 Noser} & \multicolumn{4}{c|}{A6 Toss Out} & \multicolumn{4}{c|}{A7 Cartwheel} & \multicolumn{4}{c}{AVG} \\
        \hline 
        Time (sec) & 0.2 & 0.4 & 0.6 & 1.0 & 0.2 & 0.4 & 0.6 & 1.0 & 0.2 & 0.4 & 0.6 & 1.0 & 0.2 & 0.4 & 0.6 & 1.0 & 0.2 & 0.4 & 0.6 & 1.0 & 0.2 & 0.4 & 0.6 & 1.0 & 0.2 & 0.4 & 0.6 & 1.0 & 0.2 & 0.4 & 0.6 & 1.0 \\
        \hline
        % \multirow{6}{*}{\rotatebox{90}{AME}} 
        Res-RNN~\cite{julieta2017motion} & 59 & 102 & 132 & 167 & 62 & 112 & 152 & 229 & 57 & 102 & 139 & 215 & 48 & 85 & 113 & 157 & 51 & 90 & 120 & 167 & 53 & 94 & 126 & 183 & 74 & 131 & 178 & 265 & 58 & 102 & 137 & 197 \\
        LTD~\cite{mao2019learning} & 51 & 92 & 116 & 132 & 51 & 91 & 116 & \textbf{148} & 43 & 80 & 103 & 130 & 38 & 70 & 89 & 111 & 39 & 70 & 90 & 116 & 42 & 75 & 94 & 123 & 52 & 101 & 139 & 198 & 45 & 83 & 107 & 137 \\
        HRI~\cite{mao2020history} & 34 & 69 & \underline{97} & 130 & 44 & 84 & \underline{115} & \underline{150} & 32 & 65 & 91 & 121 & 27 & 56 & 82 & 112 & 28 & 58 & 85 & 121 & 34 & 66 & 88 & 115 & 42 & 83 & 120 & 171 & 34 & 69 & 97 & 131 \\
        MSR~\cite{dang2021msr} & 41 & 75 & 99 & \underline{126} & 54 & 96 & 129 & 180 & 41 & 74 & 98 & 135 & 34 & 61 & 82 & 106 & 33 & 59 & 79 & \underline{109} & 42 & 71 & 93 & 124 & 57 & 103 & 146 & 210 & 43 & 77 & 104 & 141 \\
        XIA~\cite{guo2021multi} & \underline{32} & \underline{68} & 99 & 128 & \underline{41} & \underline{82} & 116 & 163 & \underline{29} & \underline{58} & \underline{84} & \underline{116} & \underline{24} & \textbf{50} & \textbf{73} & \textbf{96} & \textbf{24} & \underline{51} & \underline{75} & \underline{109} & \textbf{31} & \underline{62} & \underline{86} & \underline{114} & \underline{41} & \underline{81} & \underline{115} & \underline{160} & \underline{32} & \underline{65} & \underline{93} & \underline{127} \\
        Ours & \textbf{31} & \textbf{66} & \textbf{93} & \textbf{120} & \textbf{40} & \textbf{78} & \textbf{109} & \underline{150} & \textbf{27} & \textbf{54} & \textbf{77} & \textbf{109} & \textbf{23} & \textbf{50} & \underline{74} & \underline{98} & \textbf{24} & \textbf{49} & \textbf{71} & \textbf{104} & \textbf{31} & \textbf{61} & \textbf{84} & \textbf{112} & \textbf{37} & \textbf{77} & \textbf{111} & \textbf{155} & \textbf{30} & \textbf{62} & \textbf{88} & \textbf{121} \\
        % \bottomrule
        \hline
    \end{tabular}
    % \end{small}
    \end{center}
    % \vskip -0.3in
\end{table*}

\subsection{More Qualitative Results}
\label{app_subsec:qualitative}
\begin{figure*}[ht]
	\begin{center}
		\centerline{\includegraphics[width=0.7\textwidth]{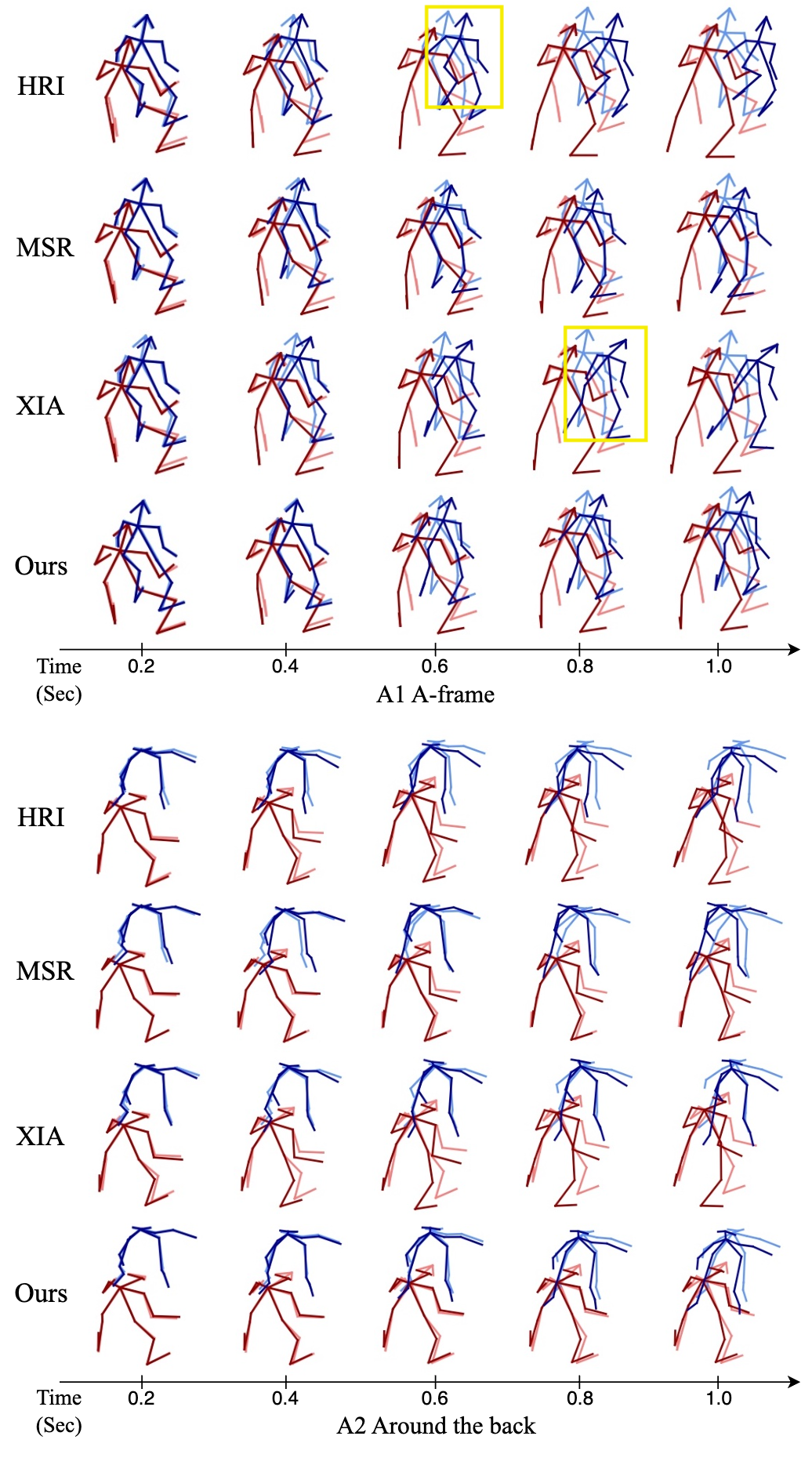}} % \linewidth
        % \vskip -0.1in
		\caption{Qualitative results of actions A1 -- A2 on the common action split. Dark red/blue represents the prediction results, while light red/blue indicates the ground truths. }
		\label{fig:visual_a1a2}
	\end{center}
%   \vskip -0.3in
\end{figure*}

\begin{figure*}[ht]
	\begin{center}
		\centerline{\includegraphics[width=0.7\textwidth]{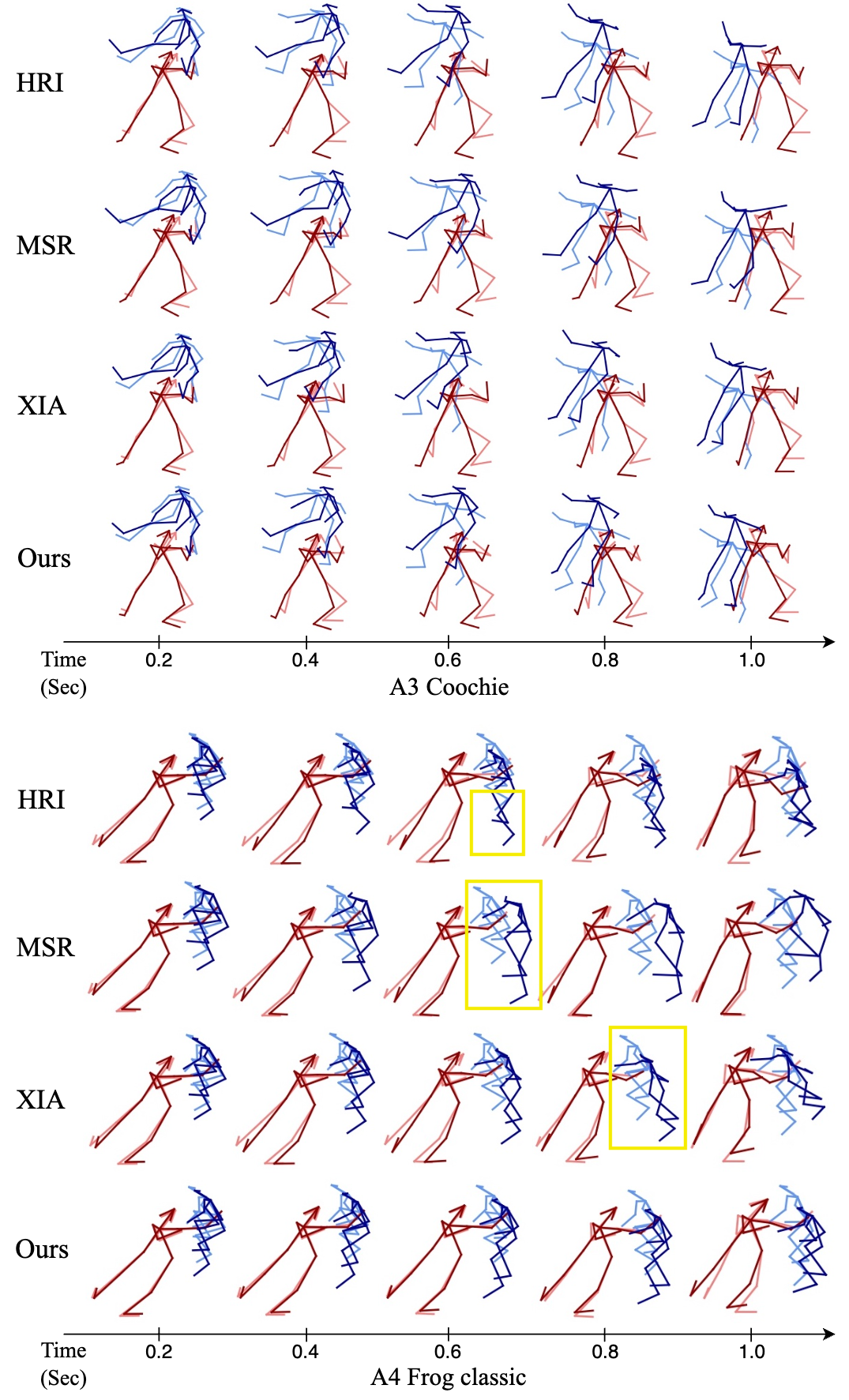}} % \linewidth
        % \vskip -0.1in
		\caption{Qualitative results of actions A3 -- A4 on the common action split. Dark red/blue represents the prediction results, while light red/blue indicates the ground truths. }
		\label{fig:visual_a3a4}
	\end{center}
%   \vskip -0.3in
\end{figure*}

\begin{figure*}[ht]
	\begin{center}
		\centerline{\includegraphics[width=0.7\textwidth]{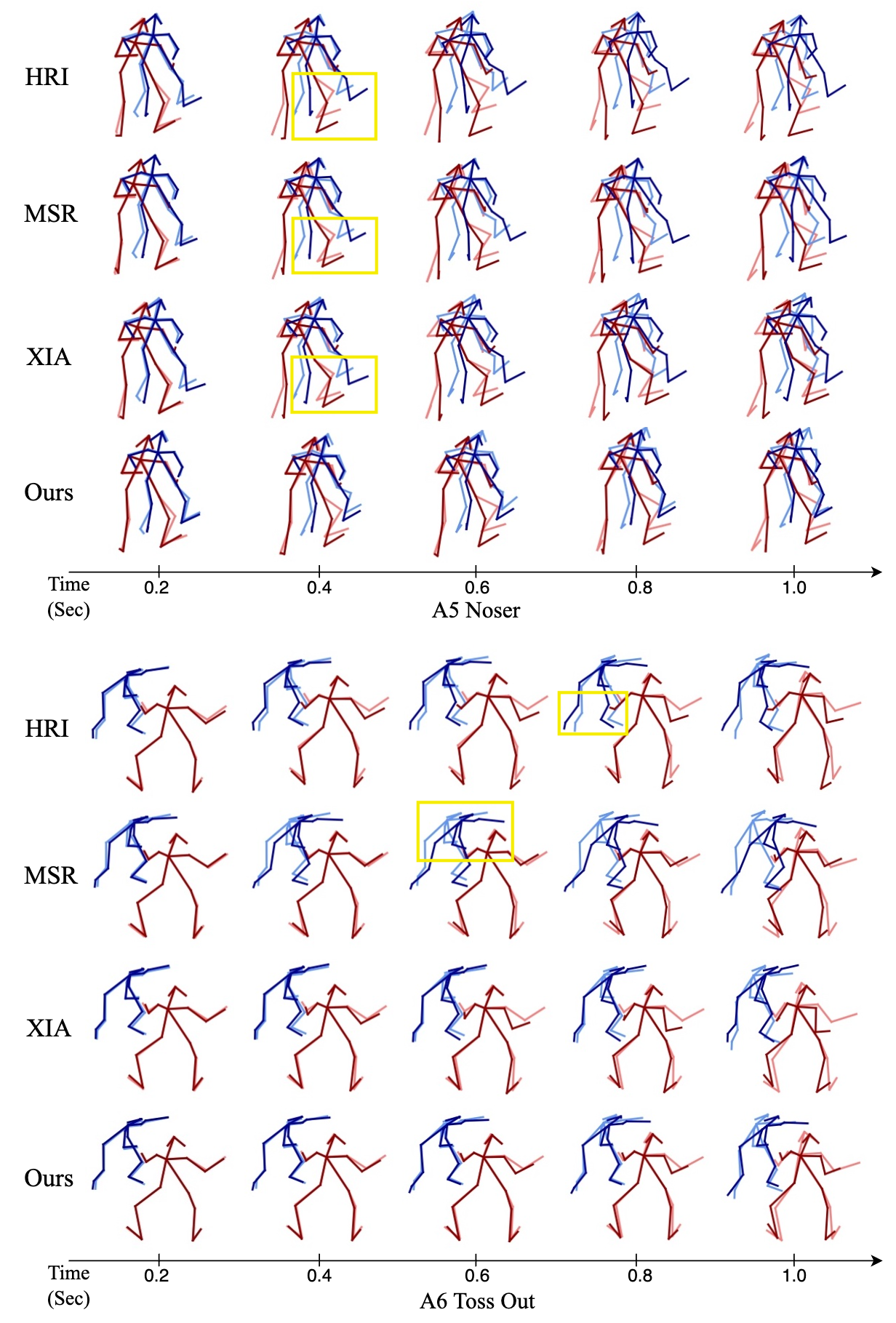}} % \linewidth
        % \vskip -0.1in
		\caption{Qualitative results of actions A5 -- A6 on the common action split. Dark red/blue represents the prediction results, while light red/blue indicates the ground truths. }
		\label{fig:visual_a5a6}
	\end{center}
%   \vskip -0.3in
\end{figure*}

\begin{figure*}[ht]
	\begin{center}
		\centerline{\includegraphics[width=0.7\textwidth]{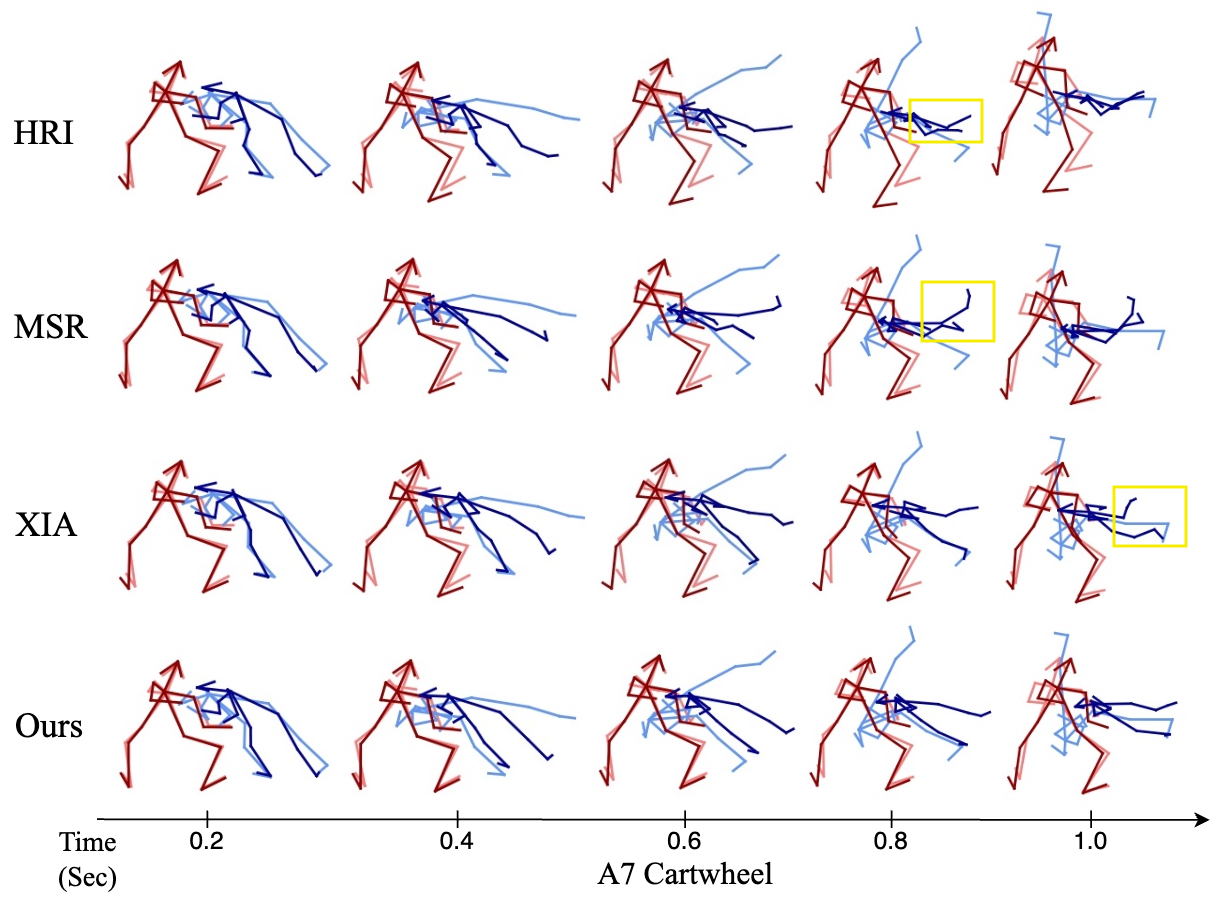}} % \linewidth
        % \vskip -0.1in
		\caption{Qualitative results of action A7 on the common action split. Dark red/blue represents the prediction results, while light red/blue indicates the ground truths. }
		\label{fig:visual_a7}
	\end{center}
%   \vskip -0.3in
\end{figure*}

More qualitative results are provided at the end of this Appendix. 
We show the examples from each action in \Cref{fig:visual_a1a2,fig:visual_a3a4,fig:visual_a5a6,fig:visual_a7}.
From these examples, with the increase of the forecasting time, the result of our PGformer becomes better than those of other compared methods that independently predict the motions of each person (HRI~\cite{mao2020history} and MSR-GCN~\cite{dang2021msr}) or only study the interactions between the historical motions (XIA-GCN~\cite{guo2021multi}).
For some extreme actions, taking A4 as an example, the poses predicted by MSR-GCN and XIA-GCN at 1 sec forecasting time are weird or look far apart from the ground truths. 
Nonetheless, our proposed PGformer successfully predicts the poses which are closer to the ground truths.

\subsection{More Ablation Study}
\label{app_subsec:ablation}
% Ablation studies are performed by using different components and hyperparameters of our network on common actions to identify their roles.
% To further identify the effects of our model's design, we ablate the main components
% Specifically, we compare different variants with our proposed one by ablating the main components, different layers and model dimensions. 
% Besides, the results also show that our proposed architecture is suitable, which has four PGformer layers with a dimension of $D=128$ and 4 heads in MHA with a dimension of 64 for each head.

We further ablate the pose encoder/decoder of our PGformer, the inner elements of our XQA module with \textit{proxy} and different hyperparameters in \Cref{tab:ablation2}. 

The variants with different pose encoding and decoding networks are first compared, and here `w/ GCN (enc)' indicates only using a GCN layer as the pose encoder while using FC layers as the pose decoder. 
Following the same spirit, our proposed model, which can be denoted as `w/ GCN (dec)', uses an FC layer as the pose encoder and GCNs as the pose decoder.
And `w/ GCN (both)' uses GCNs both in the pose encoder and decoder. 
`w/o GCN' uses FC layers instead of GCNs in the pose encoder and decoder.
For all the variants, they use a one-layer encoding network and a four-layer decoding network, which means the numbers of layers in the pose encoding and decoding network are kept the same whether FC layers or GCNs are used.
From the ablation results, we can observe that using either FC layers or GCNs as the pose encoding and decoding network has a negligible impact on the performances, but using GCNs as the pose decoder is more suitable. 
In our experiment, we also find that the pose decoding network with four layers performs better since modeling the relationships of the joints is important for the task of extreme motion prediction.

We then compare the variants with \textit{proxies} combined in different ways, and here $\bm{P}' \in \mathbb{R}^{T \times T}$  is given by: $\bm{P}' = \bm{W}_t  \bm{T} \bm{T}^{\mathsf{T}} (\bm{W}_t)^{\mathsf{T}}.$
The results show that the way in Eq. (6) influencing the bidirectional information performs the best.

Lastly, we ablate the different hyperparameters including the number of templates ($M$), number of layers and dimensions for model and FFN.
From the results, we can find that setting $M$ a small number ($M=3$ is suggested in our proposed architecture) is sufficient to build \textit{proxy}. 
From Tables {\color{red}4} and~\ref{tab:ablation2}, our suggested architecture has 4 PGformer layers in the encoder/decoder with D = 128 and $d_{ffn}$=1024 for model dimension and FFN, and 4 heads in MHA with $d_h$=64 for dimension of each head, which is simpler but more suitable.

\begin{table}[ht]
    \setlength\tabcolsep{4.0pt}
    \caption{Ablation study on the pose encoder/decoder, the inner elements of XQA module with \textit{proxy} and different hyperparameters. 
    % `w/ GCN (enc)' indicates only using a GCN layer as the pose encoder while using FC layers as the pose decoder, and `w/ GCN (both)' uses GCNs both in the pose encoder and decoder. `w/o GCN' uses FC layers instead of GCNs.
    $\otimes$ and $\oplus$ denote broadcast element-wise multiplication and addition, respectively. 
    $d_{ffn}$ is the hidden dimension of the FFN. }
    \label{tab:ablation2}
    \begin{center}
    \small
    \begin{tabular}{l|cccc|cccc}
        % \toprule
        \hline
        & \multicolumn{4}{c|}{JME} & \multicolumn{4}{c}{AME} \\
        \cline{2-9}
        Time (sec) & 0.2 & 0.4 & 0.6 & 1.0 & 0.2 & 0.4 & 0.6 & 1.0 \\
        \hline
        Proposed & \textbf{53} & \textbf{108} & \textbf{156} & \textbf{231} & \textbf{30} & \textbf{62} & \textbf{88} & \textbf{121}  \\
        \hline
        w/ GCN (enc) & 53 & 108 & 157 & 233 & 31 & 63 & 90 &  125  \\
        w/ GCN (both) & 53 & 108 & 157 & 233 & 31 & 62 & 88 & 122  \\
        w/o GCN &  53 & 109 & 158 & 234 & 30 & 62 & 88 & 123  \\
        \hline
        - $\bm{A} \otimes \bm{P}'$ & 53 & 109 & 159 & 236 & 30 & 62 & 88 & 123  \\
        - $\bm{A} \oplus \bm{P}'$ & 53 & 110 & 159 & 234 & 31 & 63 & 89 & 123  \\
        \hline
        $M = 8$ & 53 & 109 & 159 & 236 & 31 & 63 & 90 & 125 \\
        $M = 16$ & 53 & 109 & 158 & 235 & 31 & 62 & 88 & 123 \\
        3-layer & 53 & 110 & 159 & 236 & 31 & 63 & 90 & 124  \\
        6-layer & 53 & 110 & 161 & 238 & 31 & 63 & 90 & 125  \\
        % $D$=256, $H$=8, $d_h$=64 & 53 & 109 & 160 & 239 & 31 & 63 & 90 & 125 \\
        $d_{ffn}$=512 & 54 & 111 & 162 & 240 & 31 & 64 & 92 & 127 \\
        $d_{ffn}$=2048 & 54 & 110 & 161 & 237 & 31 & 63 & 91 & 126 \\
        \hline
    \end{tabular}
    \end{center}
    % \vskip -0.3in
\end{table}

\newpage
\section*{}
% \paragraph{Social Impact.}
\paragraph*{Ethics Statement.}
% Our proposed technique is useful in many applications, such as self-driving to avoid crowds. 
Our original intention for this research is to protect people’s safety in autonomous vehicles, collision avoidance for robotics and surveillance systems. 
The potential negative societal impacts include: 
(1) our approach can be used to synthesize highly realistic human motions, which might lead to the spread of false information; 
(2) there are still concerns about the invasion of people’s privacy since our approach requires real behavioral information as input, and we are concerned that this may expose the identity information. 
Nonetheless, on the positive side, our model operates on the processed human skeleton representations instead of the raw data, which contains much less identification information. 
% there are still concerns about the invasion of people’s privacy through human motion trajectories and behaviors. 

\paragraph*{Discussion of Limitations.} 
This paper mainly focuses on modeling multi-person extreme actions, while the motions from different actions vary greatly. 
Hence, it is hard to verify the effectiveness of our PGformer on other extreme actions due to the lack of such datasets. 
Besides, we only conduct the ablation study on ExPI to decide the architecture of our model. 
The performances on CMU-Mocap and MuPoTS-3D datasets would be further improved if tuning some hyperparameters.

\end{document}